\documentclass[prodmode,acmcsur]{acmsmall} 

\usepackage[ruled]{algorithm2e}

\SetAlFnt{\small}
\SetAlCapFnt{\small}
\SetAlCapNameFnt{\small}
\SetAlCapHSkip{0pt}
\IncMargin{-\parindent}

\AtBeginDocument{\setlength{\tempdimen}{\textwidth}}

\usepackage{graphicx}
\usepackage{listings}
\usepackage{url}
\usepackage{color}
\usepackage{xspace}
\usepackage{hhline}
\usepackage{multirow}
\usepackage{rotating}

\acmVolume{50}
\acmNumber{6}
 \acmArticle{84}
\acmYear{2017}
\acmMonth{11}

\newcommand{\CODMAP}{{\sffamily CoDMAP}\xspace}

\newcommand{\PDDL}{\emph{PDDL}\xspace}
\newcommand{\MAPDDL}{\emph{MA-PDDL}\xspace}
\newcommand{\MASTRIPSL}{\emph{MA-STRIPS}\xspace}

\newcommand{\NOAH}{{\sffamily NOAH}\xspace}
\newcommand{\DNOAH}{{\sffamily Distributed NOAH}\xspace}
\newcommand{\PGP}{{\sffamily PGP}\xspace}
\newcommand{\GPGP}{{\sffamily GPGP}\xspace}
\newcommand{\DSIPE}{{\sffamily DSIPE}\xspace}
\newcommand{\SIPE}{{\sffamily SIPE-2}\xspace}
\newcommand{\STEAM}{{\sffamily STEAM}\xspace}
\newcommand{\SHOP}{{\sffamily SHOP}\xspace}
\newcommand{\ASHOP}{{\sffamily A-SHOP}\xspace}
\newcommand{\MAPLAN}{{\sffamily MAPlan}\xspace}
\newcommand{\MALAPKT}{{\sffamily MAP-LAPKT}\xspace}
\newcommand{\CMAP}{{\sffamily CMAP}\xspace}
\newcommand{\MADLA}{{\sffamily MADLA}\xspace}
\newcommand{\PSM}{{\sffamily PSM}\xspace}
\newcommand{\PMR}{{\sffamily PMR}\xspace}
\newcommand{\AP}{{\sffamily A\#}\xspace}

\newcommand{\DP}{{\sffamily Distoplan}\xspace}
\newcommand{\DPGM}{{\sffamily DPGM}\xspace}
\newcommand{\MARC}{{\sffamily MARC}\xspace}
\newcommand{\MAPPOP}{{\sffamily MAP-POP}\xspace}
\newcommand{\FMAP}{{\sffamily FMAP}\xspace}
\newcommand{\MHFMAP}{{\sffamily MH-FMAP}\xspace}
\newcommand{\MAPR}{{\sffamily MAPR}\xspace}
\newcommand{\GPPP}{{\sffamily GPPP}\xspace}
\newcommand{\PF}{{\sffamily Planning First}\xspace}
\newcommand{\MASTRIPSF}{{\sffamily MA-STRIPS}\xspace}
\newcommand{\TFPOP}{{\sffamily TFPOP}\xspace}
\newcommand{\ADP}{{\sffamily ADP}\xspace}
\newcommand{\USATPLAN}{{\sffamily $\mu$-SATPLAN}\xspace}
\newcommand{\MAFS}{{\sffamily MAFS}\xspace}
\newcommand{\SMAFS}{{\sffamily Secure-MAFS}\xspace}
\newcommand{\DPP}{{\sffamily DPP}\xspace}
\newcommand{\MADA}{{\sffamily MAD-A*}\xspace}
\newcommand{\STRIPS}{{\sffamily STRIPS}\xspace}
\newcommand{\LAMA}{{\sffamily LAMA}\xspace}
\newcommand{\FF}{{\sffamily FF}\xspace}

\newcommand{\FD}{{\sffamily FD}\xspace}
\newcommand{\HSP}{{\sffamily HSP}\xspace}
\newcommand{\SATPLAN}{{\sffamily SATPLAN}\xspace}
\newcommand{\IBACOP}{{\sffamily IBACOP}\xspace}

\newcommand{\LAPKT}{{\sffamily LAPKT}\xspace}

\newcommand{\EXPLANTECH}{{\sffamily ExPlanTech}\xspace}
\newcommand{\BIOMAS}{{\sffamily BioMAS}\xspace}
\newcommand{\DECAF}{{\sffamily DECAF}\xspace}
\newcommand{\MAG}{{\sffamily Magentix2}\xspace}
\newcommand{\mands}{{\sffamily \emph{Merge\&Shrink}}\xspace}
\newcommand{\lmcut}{{\sffamily \emph{LM-Cut}}\xspace}
\definecolor{darkgreen}{rgb}{0,0.5,0}
\definecolor{lightgray}{gray}{0.92}
\definecolor{softgray}{gray}{0.96}
\definecolor{darkblue}{rgb}{.11,.23,.65}
\newcommand{\G}{{\mathcal G}}

\newcommand{\I}{{\mathcal I}}
\newcommand{\A}{{\mathcal A}}
\newcommand{\T}{{\mathcal T}}
\newcommand{\AG}{{\mathcal{AG}}}

\newcommand{\Prop}{{\mathcal P}}

\newcommand{\PRE}{pre(\alpha)}

\newcommand{\ADD}{add(\alpha)}
\newcommand{\DEL}{del(\alpha)}

\newcommand{\emphcgoal}{\emph{group goal}\xspace}
\newcommand{\cgoal}{group goal\xspace}
\newcommand{\cgoals}{group goals\xspace}
\newcommand{\ttagency}{{\ttfamily transport-agency}\xspace}
\newcommand{\ttagencies}{{\ttfamily transport-agencies}\xspace}

\doi{https://doi.org/10.1145/3128584}

\issn{1234-56789}

\begin{document}

\markboth{A. Torre\~no et al.}{Cooperative Multi-Agent Planning: A Survey}
\title{Cooperative Multi-Agent Planning: A Survey}
\author{ALEJANDRO TORRE\~NO
\affil{Universitat Polit\`ecnica de Val\`encia}
EVA ONAINDIA
\affil{Universitat Polit\`ecnica de Val\`encia}
ANTON\'IN KOMENDA
\affil{Czech Technical University in Prague}
MICHAL \v{S}TOLBA
\affil{Czech Technical University in Prague}
}

\begin{abstract}

Cooperative multi-agent planning (MAP) is a relatively recent research field that combines technologies, algorithms and techniques developed by the Artificial Intelligence Planning and Multi-Agent Systems communities. While planning has been generally treated as a single-agent task, MAP generalizes this concept by considering multiple intelligent \emph{agents} that work cooperatively to develop a course of action that satisfies the goals of the group.

This paper reviews the most relevant approaches to MAP, putting the focus on the solvers that took part in the 2015 Competition of Distributed and Multi-Agent Planning, and classifies them according to their key features and relative performance.




\end{abstract}

%
%


%
%




\acmformat{Alejandro Torre\~no, Eva Onaindia, Anton\'in Komenda, Michal \v{S}tolba, 2016. Distributed and multi-agent planning: a survey.}

\begin{bottomstuff}
This work is supported by the Spanish MINECO under project TIN2014-55637-C2-2-R, the Prometeo project II/2013/019 funded by the Valencian Government, and the 4-year FPI-UPV research scholarship granted to the first author by the Universitat Polit\`ecnica de Val\`encia. Additionally, this research is partially supported by the Czech Science Foundation under grant 15-20433Y.

Author's addresses: A. Torre\~no (atorreno@dsic.upv.es) {and} E. Onaindia (onaindia@dsic.upv.es), Universitat Polit\`ecnica de Val\`encia, Camino de Vera, s/n, Valencia, 46022, Spain; A. Komenda (antonin.komenda@fel.cvut.cz) {and} M. \v{S}tolba (michal.stolba@agents.fel.cvut.cz), Czech Technical University in Prague, Zikova 1903/4, 166 36, Prague, Czech Republic.
\end{bottomstuff}

\maketitle

\section{Introduction}
\label{introduction}

Automated Planning is the field devoted to studying the reasoning side of acting. From the restricted conceptual model assumed in classical planning to the extended models that address temporal planning, on-line planning or planning in partially-observable and non-deterministic domains, the field of Automated Planning has experienced huge advances \cite{Ghallab04}.

Multi-Agent Planning (MAP) introduces a new perspective in the resolution of a planning task with the adoption of a distributed problem-solving scheme instead of the classical single-agent planning paradigm. Distributed planning is required "when planning knowledge or responsibility is distributed among agents or when the execution capabilities that must be employed to successfully achieve objectives are inherently distributed" \cite{Desjardins99a}. 

The authors of \cite{Desjardins99a} analyze distributed planning from a twofold perspective; one approach, named \emph{Cooperative Distributed Planning}, regards a MAP task as the process of formulating or executing a plan among a number of participants; the second approach, named \emph{Negotiated Distributed Planning}, puts the focus on coordinating and scheduling the actions of multiple agents in a shared environment. The first approach has evolved to what is nowadays commonly known as \emph{cooperative and distributed MAP}, with a focus on extending planning into a distributed environment and allocating the planning task among multiple agents. The second approach is primarily concerned with controlling and coordinating the actions of multiple agents in a shared environment so as to ensure that their local objectives are met. We will refer to this second approach, which stresses the coordination and execution of large-scale multi-agent planning problems, as \emph{decentralized planning for multiple agents}. Moreover, while the first planning-oriented view of MAP relies on deterministic approaches, the study of decentralized MAP has yielded an intensive research work on coordination of activities in contexts under uncertainty and/or partial observability with the development of formal methods inspired by the use of Markov Decision Processes \cite{Seuken08}.

This paper surveys deterministic cooperative and distributed MAP methods. Our intention is to provide the reader with a broad picture of the current state of the art in this field, which has recently gained much attention within the planning community thanks to venues such as the Distributed and Multi-Agent Planning workshop\footnote{http://icaps16.icaps-conference.org/dmap.html} and the 2015 Competition of Distributed and Multi-Agent Planning\footnote{http://agents.fel.cvut.cz/codmap} (\CODMAP). Interestingly, although there was a significant amount of work on planning in multi-agent systems in the 90's, most of this research was basically aimed at developing coordination methods for agents that adopt planning representations and algorithms to carry out their tasks. Back then, little attention was given to the problem of formulating collective plans to solve a planning task. However, the recent \CODMAP initiative of fostering \MASTRIPSF  \textendash a classical planning model for multi-agent systems \cite{Brafman08}\textendash $\;$ has brought back a renewed interest.

Generally speaking, cooperative MAP is about the collective effort of multiple planning agents to develop solutions to problems that each could not have solved as well (if at all) alone \cite{Durfee99}. A cooperative MAP task is thus defined as the collective effort of multiple agents towards achieving a \emph{common goal}, irrespective of how the goals, the knowledge and the agents' abilities are distributed in the application domain. In \cite{deWeerdt09}, authors identify several phases to address a MAP task that can be interleaved depending on the characteristics of the problem, the agents and the planning model. Hence, MAP solving may require allocation of goals, formulating plans for solving goals, communicating planning choices and coordinating plans, and execution of plans. The work in \cite{deWeerdt09} is an overview of MAP devoted to agents that plan and interact, presenting a rough outline of techniques for cooperative MAP and decentralized planning. A more recent study examines how to integrate planning algorithms and Belief-Desire-Intention (BDI) agent reasoning \cite{MeneguzziS15}. This survey puts the focus on the integration of agent behaviour aimed at carrying out predefined plans that accomplish a goal and agent behaviour aimed at formulating a plan that achieves a goal. 

This paper presents a thorough analysis of the advances in cooperative and distributed MAP that have lately emerged in the field of Automated Planning. Our aim is to cover the wide and fragmented space of MAP approaches, identifying the main characteristics that define tasks and solvers and establishing a taxonomy of the main approaches of the literature. We explore the great variety of MAP techniques on the basis of different criteria, like agent distribution, communication or privacy models, among others. The survey thus offers a deep analysis of techniques and domain-independent MAP solvers from a broad perspective, without adopting any particular planning paradigm or programming language. Additionally, the contents of this paper are geared towards reviewing the broad range of MAP solvers that participated in the 2015 \CODMAP competition.

This survey is structured in five sections. Section \ref{related_work} offers a historical background on distributed planning with a special emphasis on work that has appeared over the last two decades. Section \ref{MAP_task} discusses the main modelling features of a MAP task. Section \ref{aspects} analyzes the main aspects of MAP solvers, including distribution, coordination, heuristic search and privacy. Section \ref{taxonomy} discusses and classifies the most relevant MAP solvers in the literature. Finally, section \ref{ongoing} concludes and summarizes the ongoing and future trends and research directions in MAP. 

\section{Related Work: historical background on MAP}
\label{related_work}

The large body of work on distributed MAP started jointly with an intensive research activity on multi-agent systems (MAS) at the beginning of the 90's. Motivated by the distributed nature of the problems and reasoning of MAS, decentralized MAP focused on aspects related to distributed control including activities like the decomposition and allocation of tasks to agents and utilization of resources \cite{Durfee91,Wilkins98}; reducing communication costs and constraints among agents \cite{Decker92,Wolvertond98}; or incorporating group decision making for distributed plan management in collaborative settings (group decisions for selecting a high-level task decomposition or an agent assignation to a task, group processes for plan evaluation and monitoring, etc.) \cite{GroszHK99}. From this Distributed Artificial Intelligence (DAI) standpoint, MAP is fundamentally regarded as multi-agent \emph{coordination of actions in decentralized system}s.

The inherently distributed nature of tasks and systems also fostered the appearance of techniques for \emph{cooperative formation of global plans}. In DAI, this form of MAP puts greater emphasis on reasoning, stressing the deliberative planning activities of the agents as well as how and when to coordinate such planning activities to come up with a global plan. Given the cooperative nature of the planning task, where all agents are aimed at solving a common goal, this MAP approach features a more centralized view of the planning process. Investigations in this line have yield a great variety of planning and coordination methods such as techniques to merge the local plans of the agents \cite{EphratiR94,Desjardins99b,Cox04}, heuristic techniques for agents to solve their individual sub-plans \cite{Ephrati97}, mechanisms to coordinate concurrent interacting actions \cite{BoutilierB01} or distributed constraint optimization techniques to coordinate conflicts among agents \cite{Cox09}. In this latter work, the authors propose a general framework to coordinate the activities of several agents in a common environment such as partners in a military coordination, subcontractors working on a building, or airlines operating in an alliance.

Many of the aforementioned techniques and approaches were actually used by some of the early MAP tools. \DNOAH~\cite{Corkill79} is one of the first Partial-Order Planning (POP) systems that generates gradual refinements in the space of (abstract) plans using a representation similar to the Hierarchical Task Networks (HTNs). The scheme proposed in~\cite{Corkill79} relies on a distributed conflict-solving process across various agents that are able to plan without complete or consistent planning data; the limitation of \DNOAH is the amount of information that must be exchanged between planners and the lack of robustness to communication loss or error. In the domain-specific Partial Global Planning (\PGP) method~\cite{Durfee91}, agents build their partial global view of the planning problem, and the search algorithm finds local plans that can be then coordinated to meet the goals of all the agents. Generalized PGP (\GPGP) is a domain-independent extension of \PGP \cite{Decker92,Lesser04} that separates the process of coordination from the local scheduling of activities and task selection, which enables agents to communicate more abstract and hierarchically organized information and has smaller coordination overhead. \DSIPE~\cite{Desjardins99b} is a distributed version of \SIPE~\cite{Wilkins88} closely related to the \DNOAH planner. \DSIPE proposes an efficient communication scheme among agents by creating partial views of sub-plans. The plan merging process is centralized in one agent and uses the conflict-resolution principle originally proposed in \NOAH. The authors of \cite{WeerdtBTW03} propose a plan merging technique that results in distributed plans in which agents become dependent on each other, but are able to attain their goals more efficiently. 

HTN planning has also been exploited for coordinating the plans of multiple agents~\cite{Clement99}. The attractiveness of approaches that integrate hierarchical planning in agent teams such as \STEAM~\cite{Tambe97} is that they leverage the abstraction levels of the plan hierarchies for coordinating agents, thus enhancing the efficiency and quality of coordinating the agents' plans. \ASHOP \cite{DixMNZ03} is a multi-agent version of the \SHOP HTN planner \cite{NauAIKMWY03} that implements capabilities for interacting with external agents, performs mixed symbolic/numeric computations, and makes queries to distributed, heterogeneous information sources without requiring knowledge about how and where these resources are located. Moreover, authors in \cite{KabanzaSG04} propose a distributed version of \SHOP that runs on a network of clusters through the implementation of a simple distributed backtrack search scheme.

As a whole, cooperative MAP approaches devoted to the construction of a plan that solves a common goal are determined by two factors, the underlying planning paradigm and the mechanism to coordinate the formation of the plan. The vast literature on multi-agent coordination methods is mostly concerned with the task of combining and adapting local planning representations into a global consistent solution. The adaptability of these methods to cooperative MAP is highly dependent on the particular agent distribution and the plan synthesis strategy of the MAP solver. Analyzing these aspects is precisely the aim of the following sections.

\section{Cooperative Multi-Agent Planning Tasks}
\label{MAP_task}

We define a (cooperative) MAP task as a process in which several agents that are not self-interested work together to synthesize a joint plan that solves a common goal. All agents wish thereby the goal to be reached at the end of the task execution. 

First, this section presents the formalization of the components of a cooperative MAP task. Next, we discuss the main aspects that characterize a MAP task by means of two illustrative examples. Finally, we present how to model a MAP task with \MAPDDL, a multi-agent version of the well-known Planning Domain Description Language (\PDDL) \cite{Ghallab98}.

\subsection{Formalization of a MAP Task}

Most of the cooperative MAP solvers that will be presented in this survey use a formalism that stems from \MASTRIPSF \cite{Brafman08} to a lesser or greater extent. For this reason, we will use \MASTRIPSF as the baseline model for the formalization of a MAP task. \MASTRIPSF is a minimalistic multi-agent extension of the well-known \STRIPS planning model \cite{Fikes71}, which has become the most widely-adopted formalism for describing cooperative MAP tasks.

\vspace{0.2cm}

In \MASTRIPSF, a MAP task is represented through a finite number of situations or \emph{states}. States are described by a set of \emph{atoms} or \emph{propositions}. States change via the execution of planning \emph{actions}. An action in \MASTRIPSF is defined as follows: 

\begin{definition}
\label{action}
A \textbf{planning action} is a tuple $\alpha = \langle \PRE, \ADD, \DEL\rangle$, where $\PRE$,  $\ADD$ and $\DEL$ are sets of atoms that denote the preconditions, \emph{add effects}, and \emph{delete effects} of the action, respectively. 
\end{definition}

An action $\alpha$ is executable in a state $S$ if and only if all its preconditions hold in $S$; that is, $\forall p \in \PRE$, $p \in S$. The execution of $\alpha$ in $S$ generates a state $S'$ such that $S' = S \setminus \DEL \cup \ADD$.

\begin{definition}
\label{MAP_task_def}
A \textbf{MAP task} is defined as a 5-tuple $\T=\langle \AG, \Prop, \{ \A^i \}^n_{i=1},\I,\G\rangle$ with the following components:

\begin{itemize}
	\item $\AG$ is a finite set of $n$ planning entities or agents. 
	\item $\Prop$ is a finite set of atoms or propositions.
	\item $\A^i$ is the finite set of planning actions of the agent $i \in AG$. We will denote the set of actions of $\T$ as $\A = \bigcup_{\forall i \in \AG} \A^i$.
	\item $\I \subseteq \Prop$ defines the \emph{initial state} of $\T$.
	\item $\G \subseteq \Prop$ denotes the \emph{common goal} of $\T$.	
\end{itemize}
\end{definition}

A \emph{solution plan} is an ordered set of actions whose application over the initial state $\I$ leads to a state $S_g$ that satisfies the task goals; i.e., $\G \subseteq S_g$. In \MASTRIPSF a solution plan is defined as a sequence of actions  $\Pi_g=\{\Delta, \prec\}$ that attains the task goals, where $\Delta \subseteq \A$ is a non-empty set of actions and $\prec$ is a total-order relationship among the actions of $\Delta$. However, other MAP models assume a more general definition of a plan; for example, as a set of of sequences of actions (one sequence per agent), as in \cite{Kvarnstrom11}; or as a partial-order plan \cite{Torreno12ECAI}. In the following, we will consider a solution plan as a set of partially-ordered actions. 

\vspace{0.2cm}

The action distribution model of \MASTRIPSF, introduced in Definition \ref{MAP_task_def}, classifies each atom $p \in \Prop$ as either  \emph{internal} (private) to an agent $i \in \AG$, if it is only used and affected by the actions in $\A^i$, or \emph{public} to all the agents in $\AG$. $\Prop^i_{int}$ denotes the atoms that are internal to agent $i$, while $\Prop_{pub}$ refers to the public atoms of the task. The distribution of the information of a MAP task $\T$ configures the \emph{local view} that an agent $i$ has over $\T$, $\T^i$, which is formally defined as follows:

\begin{definition}
\label{Local_view}
The \textbf{local view} of a task $\T=\langle \AG, \Prop, \A,\I,\G\rangle$ by an agent $i  \in \AG$ is defined as $\T^i =  \langle\Prop^i,\A^i,\I^i,\G\rangle$, which includes the following elements:

\begin{itemize}
	\item $\Prop^i=\Prop^i_{int}\cup\Prop_{pub}$ denotes the atoms accessible by agent $i$.
	\item $\A^i \subseteq \A$ is the set of planning actions of $i$.
	\item $\I^i \subseteq \Prop^i$ is the set of atoms of the initial state accessible by agent $i$.
	\item $\G$ denotes the common goal of the task $\T$. An agent $i$ knows all the atoms of $\G$ and it will contribute to their achievement either directly (achieving a goal $g \in \G$) or indirectly (reaching effects that help other agents achieve $g$).
\end{itemize}

\end{definition}

Note that Definition \ref{Local_view} does not specify $\G^i$, a set of individual goals of an agent $i$, because in a cooperative MAP context the common goal $\G$ is shared among all agents and it is never assigned to some particular agent (see next section for more details). 

\subsection{Characterization of a MAP Task}
\label{characterization}

This section introduces a brief example based on a logistics domain \cite{Torreno14} in order to illustrate the characteristics of a MAP task. 

\begin{figure}[t]
\centering
\includegraphics[width=10.5cm]{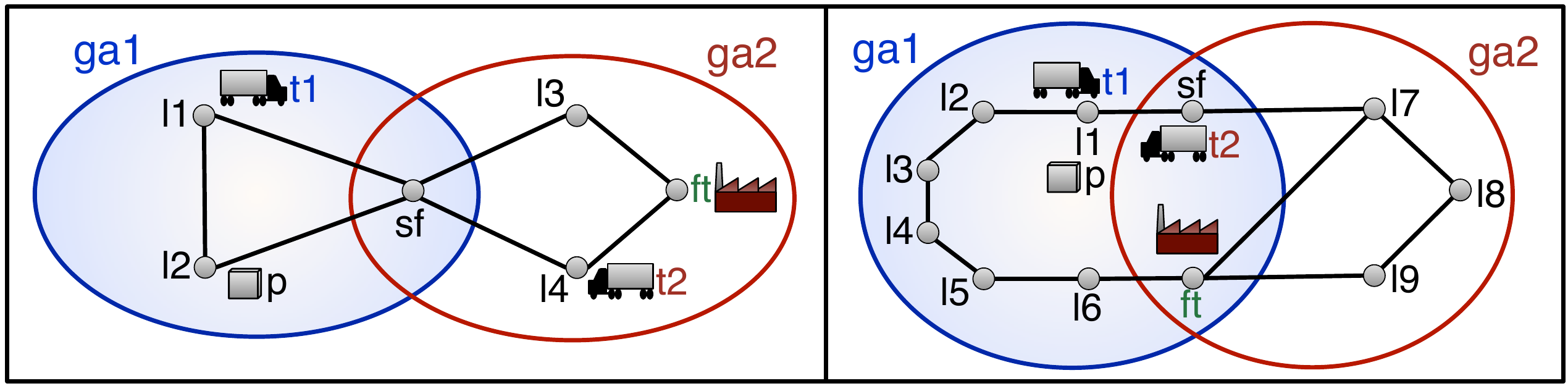}
\caption{MAP task examples: task $\T_1$ (left) and task $\T_2$ (right)}
\label{ExampleFig}
\end{figure}

Consider the transportation task $\T_1$ in Figure \ref{ExampleFig} (left), which includes three different agents. There are two transport agencies, $ta1$ and $ta2$, each of them having a truck, $t1$ and $t2$, respectively. The two transport agencies work in two different geographical areas, $ga1$ and $ga2$, respectively. The third agent is a factory, $ft$, located in the area $ga2$. In order to manufacture products, factory $ft$ requires a package of raw materials, $p$, which must be collected from area $ga1$. In this task, agents $ta1$ and $ta2$ have the same planning capabilities, but they operate in different geographical areas; i.e., they are \emph{spatially} distributed agents. Moreover, the factory agent $ft$ is \emph{functionally} different from $ta1$ and $ta2$.

The goal of task $\T_1$ is for $ft$ to manufacture a final product $fp$. For solving this task, $ta1$ will use its truck $t1$ to load the package of raw materials $p$, initially located in $l2$, and then it will transport $p$ to a storage facility, $sf$, that is located at the intersection of both geographical areas. Then, $ta2$ will complete the delivery by using its truck $t2$ to transport $p$ from $sf$ to the factory $ft$, which will in turn manufacture the final product $fp$. Therefore, this task involves three specialized agents, which are spatially or functionally distributed, and must cooperate to accomplish the common goal of having a final manufactured product $fp$. 

Task $\T_1$ defines a \emphcgoal; i.e., a goal that requires the participation of all the agents in order to solve it. Given a task $\T_g=\langle \AG, \Prop, \A, \I, \{g\}\rangle$ which includes a single goal $g$, we say that $g$ is a \cgoal if for every solution plan $\Pi_g=\{\Delta, \prec\}$, $\exists \alpha, \beta \in \Delta: \alpha \in \A^i, \beta \in \A^j$ and $i \neq j$.  We can thus distinguish between \cgoals, which require the participation of more than one agent, and non-\cgoals, which can be independently achieved by a single agent.

The presence or absence of \cgoals often determines the complexity of a cooperative MAP task. Figure \ref{ExampleFig} (right) depicts task $\T_2$, where the goal is to deliver the package $p$ into the factory $ft$. This is a non-group goal because agent $ta1$ is capable of attaining it by itself, gathering $p$ in $l1$ and transporting it to $ft$ through the locations of its area $ga1$. However, the optimal solution for this task is that agent $ta1$ takes $p$ to $sf$ so that agent $ta2$ loads then $p$ in $sf$ and completes the delivery to the factory $ft$ through location $l7$. These two examples show that cooperative MAP involves multiple agents working together to solve tasks that they are unable to attain by themselves (task $\T_1$), or tasks that are accomplished better by cooperating (task $\T_2$) \cite{deWeerdt09}.

\begin{center}
\begin{table}[h]
\bgroup
\def\arraystretch{1.2}
\centering
\tbl{Task view $\T^i$ for each agent $i$ in example tasks $\T_1$ and $\T_2$}{
{\footnotesize
\scalebox{0.92}{
\begin{tabular}{ | c || c | c | c || c | c | c |}
	\hhline{-------}
    \multicolumn{1}{|c||}{Task} &
    \multicolumn{3}{|c||}{$\T_1$} &
    \multicolumn{3}{|c|}{$\T_2$} \\ \hhline{-------}
    \multicolumn{1}{|c||}{$\AG$} &
    \multicolumn{1}{|c|}{$ta1$} &
    \multicolumn{1}{|c|}{$ta2$} &
	\multicolumn{1}{|c||}{$ft$} &    
    \multicolumn{1}{|c|}{$ta1$} &
    \multicolumn{1}{|c|}{$ta2$} &
	\multicolumn{1}{|c|}{$ft$} \\ \hhline{=======}
	\multicolumn{1}{|c||}{\multirow{3}{*}{$\Prop^i$}} &
    \multicolumn{1}{|c|}{$\mathit{(pos\;t1\;*)}$} &
    \multicolumn{1}{|c|}{$\mathit{(pos\;t2\;*)}$} &
	\multicolumn{1}{|c||}{$\mathit{(pending\;fp)}$} &    
    \multicolumn{1}{|c|}{$\mathit{(pos\;t1\;*)}$} &
    \multicolumn{1}{|c|}{$\mathit{(pos\;t2\;*)}$} &
  	\multicolumn{1}{|c|}{$\mathit{(pending\;fp)}$} \\ \cline{2-7}
	 & \multicolumn{2}{|c|}{$\mathit{(at\;p\;*)}$} & $\mathit{(at\;p\;ft)}$ & \multicolumn{2}{|c|}{$\mathit{(at\;p\;*)}$} & $\mathit{(at\;p\;ft)}$ \\ \cline{2-7}
	 & \multicolumn{3}{|c||}{$\mathit{(manufactured\;fp)}$} & \multicolumn{3}{|c|}{$\mathit{(manufactured\;fp)}$} \\
	\hhline{=======}
	\multicolumn{1}{|c||}{\multirow{1}{*}{$\A^i$}} &
    \multicolumn{2}{|c|}{drive, load, unload} & manufacture & \multicolumn{2}{|c|}{drive, load, unload} & manufacture \\ \hhline{=======}

    \multicolumn{1}{|c||}{\multirow{3}{*}{$\I^i$}} & 
    $\mathit{(pos\;t1\;l1)}$ & & & $\mathit{(pos\;t1\;l1)}$ & & \\
	&  & $\mathit{(pos\;t2\;l4)}$ & $\mathit{(pending\;fp)}$ & $\mathit{(at\;p\;l1)}$ &  $\mathit{(pos\;t2\;sf)}$ & \\
	& $\mathit{(at\;p\;l2)}$ &  & & $\mathit{(at\;p3\;ft)}$ &  &  \\ \hhline{=======}

    \multicolumn{1}{|c||}{$\G$} & \multicolumn{3}{|c||}{$\mathit{(manufactured\;fp)}$} & \multicolumn{3}{|c|}{$\mathit{(at\;p\;ft)}$} \\
	\hhline{-------}
\end{tabular}}}
}
\egroup
\label{views}
\end{table}
\end{center}

Tasks $\T_1$ and $\T_2$ emphasize most of the key elements of a MAP context. The spatial and/or functional distribution of the participants gives rise to \emph{specialized agents} that have different capabilities and knowledge of the task. Information of the MAP tasks is distributed among the specialized agents as summarized in Table \ref{views}. Atoms of the form $\mathit{(pos\;t1\;*)}$ (note that $*$ acts as a wildcard) are accessible to agent $ta1$, since they model the position of truck $t1$. Atoms of the form $\mathit{(pos\;t2\;*)}$, which describe the location of truck $t2$, are accessible to agent $ta2$. Finally, $\mathit{(pending\;fp)}$ belongs to agent $ft$ and denotes that the manufacturing of $fp$ (the goal of task $\T_1$) is still pending.

The atoms related to the location of the product $p$, $\mathit{(at\;p\;*)}$, as well as $\mathit{(manufactured\;fp)}$, which indicates that the final product $fp$ is already manufactured, are accessible to the three agents, $ta1$, $ta2$ and $ft$. Since agents ignore the configuration of the working area of the other agents, the knowledge of agent $ta1$ regarding the location of $p$ is restricted to the atoms $\mathit{(at\;p\;l1)}$, $\mathit{(at\;p\;l2)}$, $\mathit{(at\;p\;sf)}$ and $\mathit{(at\;p\;t1)}$, while agent $ta2$ knows $\mathit{(at\;p\;sf)}$, $\mathit{(at\;p\;l3)}$, $\mathit{(at\;p\;l4)}$, $\mathit{(at\;p\;t2)}$ and $\mathit{(at\;p\;ft)}$. The awareness of agent $ft$ with respect to the location of $p$ is limited to $\mathit{(at\;p\;ft)}$. 

\vspace{0,2cm}

The information distribution of a MAP task stresses the issue of \emph{privacy}, which is one of the basic aspects that must be considered in multi-agent applications \cite{Such13}. Agents manage information that is not relevant for their counterparts or sensitive data of their internal operational mechanisms that they are not willing to disclose. For instance, $ta1$ and $ta2$ cooperate in solving tasks $\T_1$ and $\T_2$ but they could also be potential competitors since they work in the same business sector. For these reasons, providing privacy mechanisms to guarantee that agents do not reveal the internal configuration of their working areas to each other is a key issue. 

In general, agents in MAP seek to minimize the information they share with each other, thus exchanging only the information that is relevant for other participating agents to solve the MAP task. 

\subsection{Modelling of a MAP Task with \MAPDDL}
\label{specification}

The adoption of a common language for modelling planning domains allows for a direct comparison of different approaches and increases the availability of shared planning resources, thus facilitating the scientific development of the field \cite{Fox03}. Modelling a cooperative MAP task involves defining several elements that are not present in single-agent planning tasks. Widely-adopted single-agent planning task specification languages, such as \PDDL \cite{Ghallab98}, lack the required machinery to specify a MAP task. Recently, \MAPDDL\footnote{Please refer to \url{http://agents.fel.cvut.cz/codmap/MA-PDDL-BNF-20150221.pdf} for a complete BNF definition of the syntax of \MAPDDL.}, the multi-agent version of \PDDL, was developed in the context of the 2015 \CODMAP competition \cite{Komenda16codmap} as the first attempt to create a \emph{de facto} standard specification language for MAP tasks. We will use \MAPDDL as the language for modelling MAP tasks.

\vspace{0.2cm}

MAP solvers that accept an \emph{unfactored} specification of a MAP task use a single input that describes the complete task $\T$. In contrast, other MAP approaches require a \emph{factored} specification; i.e., the local view of each agent, $\T^i$. Additionally, modelling a MAP task may require the specification of the private information that an agent cannot share with other agents. 

\MAPDDL allows for the definition of both factored ({\ttfamily :factored-privacy} requirement) and unfactored ({\ttfamily :unfactored-privacy} requirement) task representations. In order to model the transportation task $\T_1$ (see Figure \ref{ExampleFig} (left) in Section \ref{characterization}), we will use the factored specification. Task $\T^i_1$ of agent $i$ is encoded by means of two independent files: the \emph{domain} file describes general aspects of the task ($\Prop^i$ and $\A^i$, which can be reused for solving other tasks of the same typology); the \emph{problem} file contains a description of the particular aspects of the task to solve ($\I^i$ and $\G$). For the sake of simplicity, we only display fragments of the task $\T_1^{ta1}$.

\vspace{0.2cm}

The domain description of agents of type \ttagency, like $ta1$ and $ta2$, is defined in Listing \ref{domain_agency}.
 
{\ttfamily\begin{lstlisting}[label=domain_agency, caption=Excerpt of the domain file for \ttagency agents]
(define (domain transport-agency)
  (:requirements :factored-privacy :typing :equality :fluents)    
  (:types transport-agency area location package product - object
          truck place - location
          factory - place)  
  (:predicates 
    (manufactured ?p - product) (at ?p - package ?l - location)
    (:private
      (area ?ag - transport-agency ?a - area) (in-area ?p - place ?a - area)
      (owner ?a - transport-agency ?t - truck) (pos ?t - truck ?l - location)
      (link ?p1 - place ?p2 - place) 
    )
  )  
  (:action drive
    :parameters   (?ag - transport-agency ?a - area ?t - truck ?p1 - place ?p2 - place)
    :precondition (and (area ?ag ?a) (in-area ?p1 ?a) (in-area ?p2 ?a)
                  (owner ?a ?t) (pos ?t ?p1) (link ?p1 ?p2))
    :effect       (and (not (pos ?t ?p1)) (pos ?t ?p2))
  )  
[...]
)
\end{lstlisting}} 

The domain of \ttagencies starts with the type hierarchy, which includes the types \ttagency and {\ttfamily factory} to define the agents of the task. Note that the type {\ttfamily factory} is defined as a subtype of {\ttfamily place} because a {\ttfamily factory} is also interpreted as a {\ttfamily place} reachable by a truck.  The remaining elements of the task are identified by means of the types {\ttfamily location}, {\ttfamily package}, {\ttfamily truck}, etc. 

The {\ttfamily :predicates} section includes the set of first-order \emph{predicates}, which are patterns to generate the agent's propositions, $\Prop^i$, through the instantiation of their parameters. The domain for \ttagencies includes the public predicates {\ttfamily at}, which models the position of the {\ttfamily packages}, and {\ttfamily manufactured}, which indicates that the task goal of manufacturing a {\ttfamily product} is fulfilled. Despite the fact that only the {\ttfamily factory} agent $ft$ has the ability to manufacture {\ttfamily products}, all the agents have access to the predicate {\ttfamily manufactured} so that the \ttagencies will be informed when the task goal is achieved.

The remaining predicates in Listing \ref{domain_agency} are in the section {\ttfamily :private} of each \ttagency, meaning that agents will not disclose information concerning the topology of their working {\ttfamily areas} or the status of their {\ttfamily trucks}.

Finally, the {\ttfamily :action} block of Listing \ref{domain_agency} shows the \emph{action schema} {\ttfamily drive}. An action schema represents a number of different actions that can be derived by instantiating its variables. Agents $ta1$ and $ta2$ have three action schemas: {\ttfamily load}, {\ttfamily unload} and {\ttfamily drive}. 

\vspace{0.2cm}

Regarding the \emph{problem} description, the factored specification includes three problem files, one per agent, that contain $\I^i$, the initial state of each agent, and the task goals $\G$.

{\ttfamily\begin{lstlisting}[label=problem_ta1, caption=Problem file of agent $ta1$]
(define (problem ta1)
  (:domain transport-agency)
  (:objects
    ta1 - transport-agency
    ga1 - area                  l1 l2 sf - place
    p - package                 fp - product
    (:private t1 - truck)
  )    
  (:init
    (area ta1 ga1) (pos t1 l1) (owner t1 ta1) (at p l1)
    (link l1 l2) (link l2 l1) (link l1 sf)
    (link sf l1) (link l2 sf) (link sf l2)
    (in-area l1 ga1) (in-area l2 ga1) (in-area sf ga1)
  )  
  (:goal (manufactured fp))
)
\end{lstlisting}}

Listing \ref{problem_ta1} depicts the problem description of task $\T_1^{ta1}$. The information managed by $ta1$ is related to its {\ttfamily truck t1} (which is defined as {\ttfamily :private} to prevent $ta1$ from disclosing the location or cargo of {\ttfamily t1}), along with the {\ttfamily places} within its working {\ttfamily area}, {\ttfamily ga1}, and the {\ttfamily package p}. The {\ttfamily :init} section of Listing \ref{problem_ta1} specifies $\I^{ta1}$; i.e., the location of {\ttfamily truck t1} and the position of {\ttfamily package p}, which is initially located in {\ttfamily ga1}. Additionally, $ta1$ is aware of the {\ttfamily links} and {\ttfamily places} within its {\ttfamily area} {\ttfamily ga1}. The {\ttfamily :goal} section is common to the three agents in $\T_1$ and includes a single goal indicating that the {\ttfamily product fp} must be {\ttfamily manufactured}.

\vspace{0,2cm}

This modelling example shows the flexibility of \MAPDDL for encoding the specific requirements of a MAP task, such as the agents' distributed information via factored input and the private aspects of the task. These functionalities make \MAPDDL a fairly expressive language to specify MAP tasks.

\section{Main Aspects of a MAP Solver}
\label{aspects}

Solving a cooperative MAP task requires various features such as information distribution, specialized agents, coordination or privacy. The different MAP solving techniques in the literature can be classified according to the mechanisms they use to address these functionalities. We identify six main features to categorize cooperative MAP solvers:

\begin{itemize}
	\item \textbf{Agent distribution}: From a conceptual point of view, MAP is regarded as a task in which multiple agents are involved, either as entities participating in the plan synthesis (planning agents) or as the target entities of the planning process (actuators or execution agents).
		
	\item \textbf{Computational process}: From a computational perspective, MAP solvers use a \emph{centralized} or \emph{monolithic} design that solves the MAP task through a central process, or a \emph{distributed} approach that splits the planning activity among several processing units.

	\item \textbf{Plan synthesis schemes}: There exist a great variety of strategies to tackle the process of synthesizing a plan for the MAP task, mostly characterized by how and when the \emph{coordination} activity is applied. Coordination comprises the distributed information exchange processes by which the participating agents organize and harmonize their activities in order to work together properly.
	
	\item \textbf{Communication mechanisms}: Communication among agents is an essential aspect that distinguishes MAP from single-agent planning. The type of communication enabled in MAP solvers is highly dependent on the type of computational process (centralized or distributed) of the solver. Thus, we will classify solvers according to the use of internal or external communication infrastructures.

	\item \textbf{Heuristic search}: As in single agent planning, MAP solvers commonly apply heuristic search to guide the planning process. In MAP, we can distinguish between \emph{local} heuristics (each agent $i$ calculates an estimate to reach the task goals $\G$ using only its accessible information in $\T^i$) or \emph{global} heuristics (the estimate to reach $\G$ is calculated among all the agents in $\AG$).
	
	\item \textbf{Privacy preservation}: Privacy is one of the main motivations to adopt a MAP approach. Privacy means coordinating agents without making sensitive information publicly available. Whereas this aspect was initially neglected in former MAP solvers \cite{Krogt09}, the most recent approaches tackle this issue through the development of robust privacy-preserving algorithms.

\end{itemize}

The following subsections provide an in-depth analysis of these aspects, which characterize and determine the performance of the existing MAP solvers.

\subsection{Agent Distribution}

Agents in MAP can adopt different roles: planning agents are \emph{reasoning} entities that synthesize the course of action or plan that will be later executed by a set of \emph{actuators} or \emph{execution agents}. An execution agent can be, among others, a robot in a multi-robot system, or a software entity in an execution simulator.

From a conceptual point of view, MAP solvers are characterized by the agent distribution they apply; i.e., the number of planning and execution agents involved in the task. Typically, it is assumed that one planning agent from the set $\AG$ of a MAP task $\T$ is associated with one actuator in charge of executing the actions of this planning agent in the solution plan. However, some MAP solvers alter this balance between planning and acting agents.

Table \ref{conceptual_schemes} summarizes the different schemes according to the relation between the number of planning and execution agents. Single-agent planning is the simplest mapping: the task is solved by a single planning agent, i.e., $|\AG|=1$, and executed by a single actuator. We can mention Fast Downward (\FD) \cite{Helmert06} as one of the most utilized single-agent planners within the planning community.

\begin{center}
\begin{table}[h]
\bgroup
\def\arraystretch{1.2}
\centering
\tbl{Conceptual schemes according to the number of planning and execution agents (along with example MAP solvers that apply them)}{
{\footnotesize
\begin{tabular}{| c | c || c | c |}
   \cline{3-4}
   \multicolumn{2}{c|}{} & \multicolumn{2}{c|}{Planning agents $|\AG|$} \\ \cline{3-4}
   \multicolumn{2}{c|}{} & $1$ & $n$ \\ \hhline{--==}
   \multirow{4}{*}{Execution agents} & \multirow{2}{*}{$1$} & Single-agent planning & Factored planning \\
   &  & \FD \cite{Helmert06} & \ADP \cite{Crosby13} \\ \cline{2-4}
   & \multirow{2}{*}{$n$} & Planning \emph{for} multiple agents & Planning \emph{by} multiple agents \\
   & & \TFPOP \cite{Kvarnstrom11} & \FMAP \cite{Torreno15} \\ \hline

\end{tabular}}
}
\egroup
\label{conceptual_schemes}
\end{table}
\end{center}

MAP solvers like \DP \cite{Fabre10}, \AP \cite{Jezequel12} and \ADP \cite{Crosby13} follow a \emph{factored} agent distribution inspired by the factored planning scheme \cite{Amir03}. Under this paradigm, a single-agent planning task is decomposed into a set of independent factors (agents), thus giving rise to a MAP task with $|\AG|>1$. Then, factored methods to solve the agents' local tasks $\T^i$ are applied, and finally, the computed local plans are pieced together into a valid solution plan \cite{Brafman06}. Factored planning exploits locality of the solutions and a limited information propagation between components.

The second row of Table \ref{conceptual_schemes} outlines the classification of MAP approaches that build a plan that is conceived to be then executed by several actuators. Some solvers in the literature regard MAP as a single planning agent working \emph{for} a set of actuators ($|\AG|=1$), while other approaches regard MAP as planning \emph{by} multiple planners  ($|\AG|>1$).

\subsubsection{Planning \emph{for} multiple agents}

Under this scheme, the actions of the solution plan are distributed among actuators typically via the introduction of constraints. \TFPOP \cite{Kvarnstrom11} applies single-agent planning for multi-agent domains where each execution agent is associated with a sequential thread of actions within a partial-order plan. The combination of forward-chaining and least-commitment of \TFPOP provides flexible schedules for the acting agents, which execute their actions in parallel. The work in \cite{Crosby14} transforms a MAP task that involves multiple agents acting concurrently and cooperatively in a single-agent planning task. The transformation compels agents to select joint actions associated with a single subset of objects at a time, and ensures that the concurrency constraints on this subset are satisfied.The result is a single-agent planning problem in which agents perform actions individually, one at a time.

The main limitation of this planning-for-multiple-agents scheme is its lack of privacy, since the planning entity has complete access to the MAP task $\T$. This is rather unrealistic if the agents involved in the task have sensitive private information they are not willing to disclose \cite{Sapena08}. Therefore, this scheme is not a suitable solution for privacy-preserving MAP tasks like task $\T_1$ described in section \ref{characterization}.

\subsubsection{Planning \emph{by} multiple agents} This scheme distributes the MAP task among several planning agents where each is associated with a local task $\T^i$. Thus, planning-by-multiple-agents puts the focus on the coordination of the planning activities of the agents. Unlike single-planner approaches, the planning decentralization inherent to this scheme makes it possible to effectively preserve the agents' privacy.

In general, solvers that follow this scheme, such as \FMAP \cite{Torreno12ECAI}, maintain a one-to-one correspondence between planning and execution agents; that is, planning agents are assumed to solve their tasks, which will be later executed by their corresponding actuators. There exist, however, some exceptions in the literature that break this one-to-one correspondence such as \MARC \cite{marc-codmap15}, which rearranges the $n$ planning agents in $\AG$ into $m$ \emph{transformer agents} ($m < n$), where a transformer agent comprises the planning tasks of several agents in $\AG$. All in all, \MARC considers $m$ reasoning entities that plan for $n$ actuators, where $m < n$.

\begin{figure}[t]
\centering
\includegraphics[width=10.4cm]{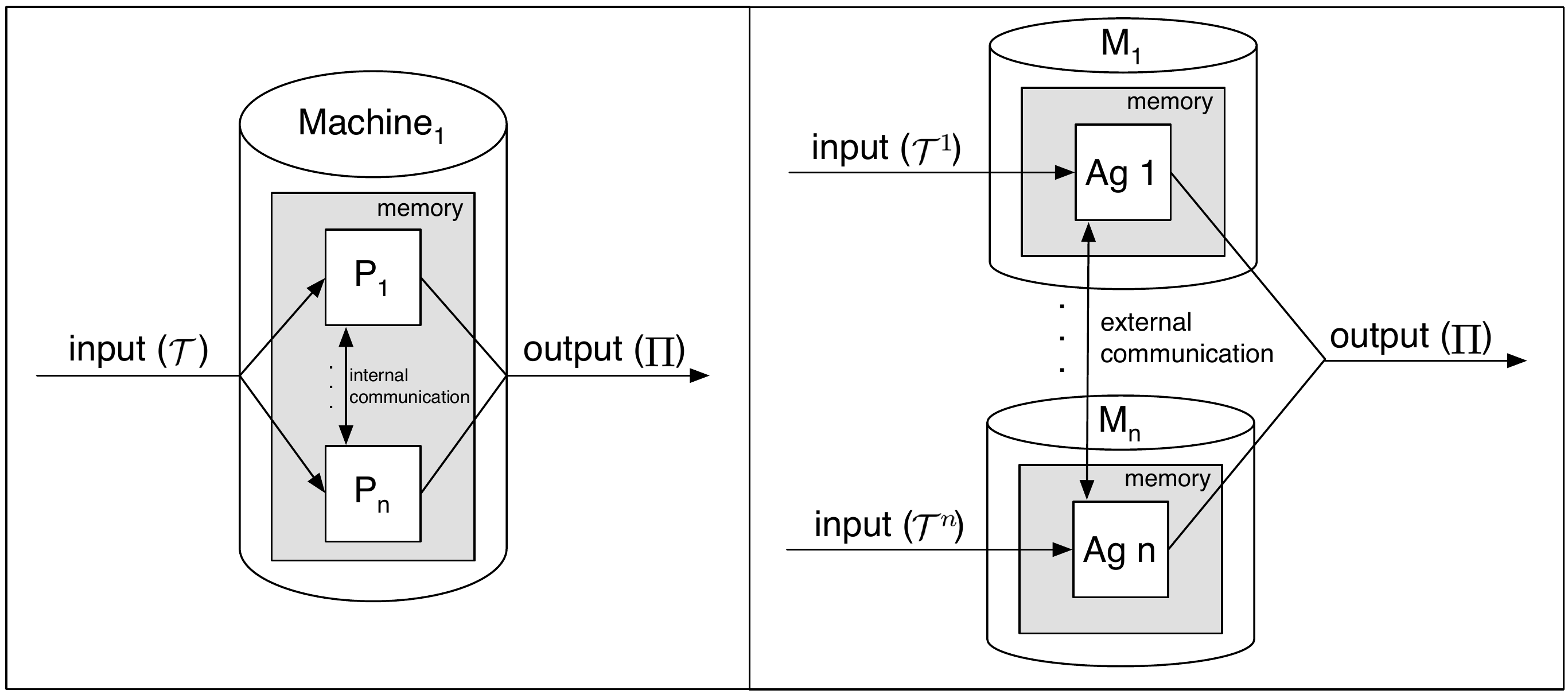}
\caption{Centralized (monolithic) vs. distributed (agent-based) implementation}
\label{FigDistribution}
\end{figure}

\subsection{Computational process}

From a computational standpoint, MAP solvers are classified as \emph{centralized} or \emph{distributed}. Centralized solvers draw upon a monolithic implementation in which a central process synthesizes a global solution plan for the MAP task. In contrast, distributed MAP methods are implemented as multi-agent systems in which the problem-solving activity is fully decentralized.

\paragraph{Centralized MAP}In this approach, the MAP task $\T$ is solved on a single machine regardless the number of planning agents conceptually considered by the solver. The main characteristic of a centralized MAP approach is that tasks are solved in a monolithic fashion, so that all the processes of the MAP solver, $\{P_1,\ldots,P_n\}$, are run in one same machine (see Figure \ref{FigDistribution} (left)).

The motivation for choosing a centralized MAP scheme is twofold: 1) external communication mechanisms to coordinate the planning agents are not needed; and 2) robust and efficient single-agent planning technology can be easily reused.

Regarding agent distribution, MAP solvers that use a single planning agent generally apply a centralized computational scheme, as for example \TFPOP \cite{Kvarnstrom11} (see Table \ref{conceptual_schemes}). On the other hand, some algorithms that conceptually rely on the distribution of the MAP task among several planning agents do not actually implement them as software agents, but as a centralized procedure. For example, \MAPR \cite{Borrajo13} establishes a sequential order among the planning agents and applies a centralized planning process that incrementally synthesizes a solution plan by solving the agents' local tasks in the predefined order.

\paragraph{Distributed MAP} Many approaches that conceive MAP as planning by multiple agents (see Table \ref{conceptual_schemes}), are developed as multi-agent systems (MAS) defined by several independent \emph{software agents}. By software agent, we refer to a computer system that 1) makes decisions without any external intervention (autonomy), 2) responds to changes in the environment (reactivity), 3) exhibits goal-directed behaviour by taking the initiative (pro-activeness), and 4) interacts with other agents via some communication language in order to achieve its objectives (social ability) \cite{Wooldridge97}.

In this context, a software agent of a MAS plays the role of a planning agent in $\AG$. This way, in approaches that follow the \emph{planning by multiple agents} scheme introduced in the previous section, a software agent encapsulates the local task $\T^i$ of a planning agent $i \in \AG$. Given a task, where $|\AG|=n$, distributed MAP solvers can be run on up to $n$ different hosts or machines (see Figure \ref{FigDistribution} (right)).

The emphasis of the distributed or agent-based computation lies in the coordination of the concurrent activities of the software planning agents. Since agents may be run on different hosts (see Figure \ref{FigDistribution} (right)), having a proper communication infrastructure and message-passing protocols is vital for the synchronization of the agents.

Distributed solvers like \FMAP \cite{Torreno14} launch $|\AG|$ software agents that seamlessly operate on different machines. \FMAP builds upon the MAS platform \MAG \cite{Such12}, which provides the messaging infrastructure for agents to communicate over the network.

\subsection{Plan synthesis schemes}
\label{planning-coordination}

In most MAP tasks, there are dependencies between agents' actions and none of the participants has sufficient competence, resources or information to solve the entire problem. For this reason, agents must coordinate with each other in order to cooperate and solve the MAP task properly. 

Coordination is a multi-agent process that harmonizes the agents' activities, allowing them to work together in an organized way. In general, coordination involves a large variety of activities, such as distributing the task goals among the agents, making joint decisions concerning the search for a solution plan, or combining the agents' individual plans into a solution for a MAP task. Since coordination is an inherently distributed mechanism, it is only required in MAP solvers that conceptually draw upon multiple planning agents (see right column of Table \ref{conceptual_schemes}). 

The characteristics of a MAP task often determine the coordination requirements for solving the task. For instance, tasks that feature \cgoals, like the task $\T_1$ depicted in section \ref{characterization}, usually demand a stronger coordination effort. Therefore, the capability and efficiency of a MAP solver is determined by the coordination strategy that governs its behaviour. 

The following subsections analyse the two principal coordination strategies in MAP; namely, \emph{unthreaded} and \emph{interleaved} planning and coordination.

\begin{figure}[t]
\centering
\includegraphics[width=13.8cm]{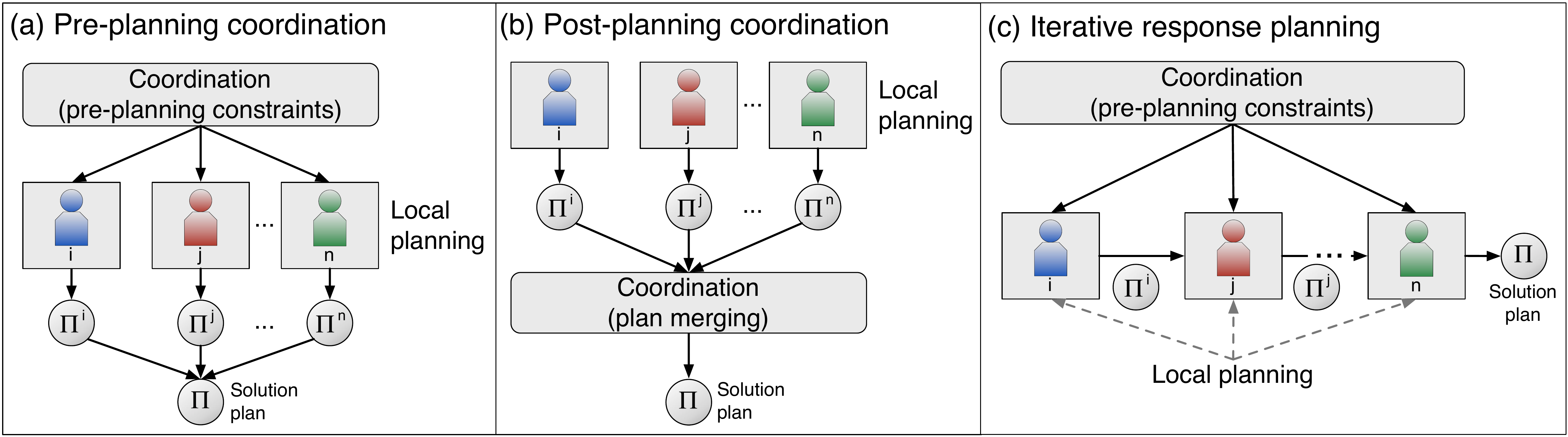}
\caption{Plan synthesis schemes in unthreaded planning and coordination}
\label{FigUnthreaded}
\end{figure}

\subsubsection{Unthreaded Planning and Coordination}

This strategy defines planning and coordination as \emph{sequential} activities, such that they are viewed as two separate black boxes. Under this strategy, an agent  $i \in \AG$ synthesizes a plan to its local view of the task, $\T^i$, and coordination takes place \emph{before} or/and \emph{after} planning. 

\paragraph{Pre-planning coordination} Under this plan synthesis scheme, the MAP solver defines the necessary constrains to guarantee that the plans that solve the local tasks of the agents are properly combined into a consistent solution plan for the whole task $\T$ (see Figure \ref{FigUnthreaded} (a)). 

\ADP \cite{Crosby13} follows this scheme by applying an \emph{agentification} procedure that distributes a \STRIPS planning task among several planning agents (see Table \ref{conceptual_schemes}). More precisely, \ADP is a fully automated process that inspects the multi-agent nature of the planning task and calculates an agent decomposition that results in a set of $n$ decoupled local tasks. By leveraging this agent decomposition, \ADP applies a centralized, sequential and total-order planning algorithm that yields a solution for the original \STRIPS task. Since the task $\T$ is broken down into several local tasks independent from each other, the local solution plans are consistently combined into a solution for $\T$.

Ultimately, the purpose of pre-planning coordination is to guarantee that the agents' local plans are seamlessly combined into a solution plan that attains the goals of the MAP task, thus avoiding the use of plan merging techniques at post-planning time.

\paragraph{Post-planning coordination} Other unthreaded MAP solvers put the coordination emphasis \emph{after} planning. In this case, the objective is to \emph{merge} the plans that solve the agents' local tasks, $\{\T^1, \ldots, \T^n\}$, into a solution plan that attains the goals $\G$ of the task $\T$ by removing  inconsistencies among the local solutions (see Figure \ref{FigUnthreaded} (b)).

In \PMR (Plan Merger by Reuse) \cite{Luis14}, the local plans of the agents are concatenated into a solution plan for the MAP task. Other post-planning coordination approaches apply an information exchange between agents to come up with the global solution. For instance, \PSM \cite{Tozicka15} draws upon a set of finite automata, called Planning State Machines (PSM), where each automaton represents the set of local plans of a given agent. In one iteration of the \PSM procedure, each agent $i$ generates a plan that solves its local task $\T^i$, incorporating this plan into its associated PSM. Then, agents exchange the public projection of their PSMs, until a solution plan for the task $\T$ is found.

\vspace{0.2cm}

\paragraph{Iterative response planning} This plan synthesis scheme, firstly introduced by \DPGM \cite{Pellier10}, successively applies a planning-coordination sequence, each corresponding to a planning agent. An agent $i$ receives the local plan of the preceding agent along with a set of constraints for coordination purposes, and \emph{responds} by building up a solution for its local task $\T^i$ on top of the received plan. Hence, the solution plan is incrementally synthesized (see Figure \ref{FigUnthreaded} (c)).

Multi-Agent Planning by Reuse (\MAPR) \cite{Borrajo13} is an iterative response solver based on \emph{goal allocation}. The task goals $\G$ are distributed among the agents before planning, such that an agent $i$ is assigned a subset $\G^i \subset \G$, where $\bigcap_{i \in \AG} \G^i = \emptyset$. Additionally, agents are automatically arranged in a sequence that defines the order in which the iterative response scheme must be carried out. 

\vspace{0.2cm}

In unthreaded planning and coordination schemes, agents do not need communication skills because they do not interact with each other during planning. This is the reason why the unthreaded strategy is particularly efficient for solving tasks that do not require a high coordination effort. In contrast, it presents several limitations when solving tasks with \cgoals, due to the fact that agents are unable to discover and address the cooperation demands of other agents. The needs of cooperation that arise when solving \cgoals are hard to discover at pre-planning time, and plan merging techniques are designed only to fix inconsistencies among local plans, rather than repairing the plans to satisfy the inter-agent coordination needs. Consequently, unthreaded approaches are more suitable for solving MAP task that do not contain \cgoals; that is, every task goal can be solved by at least one single agent.

\begin{figure}[t]
\centering
\includegraphics[width=6.2cm]{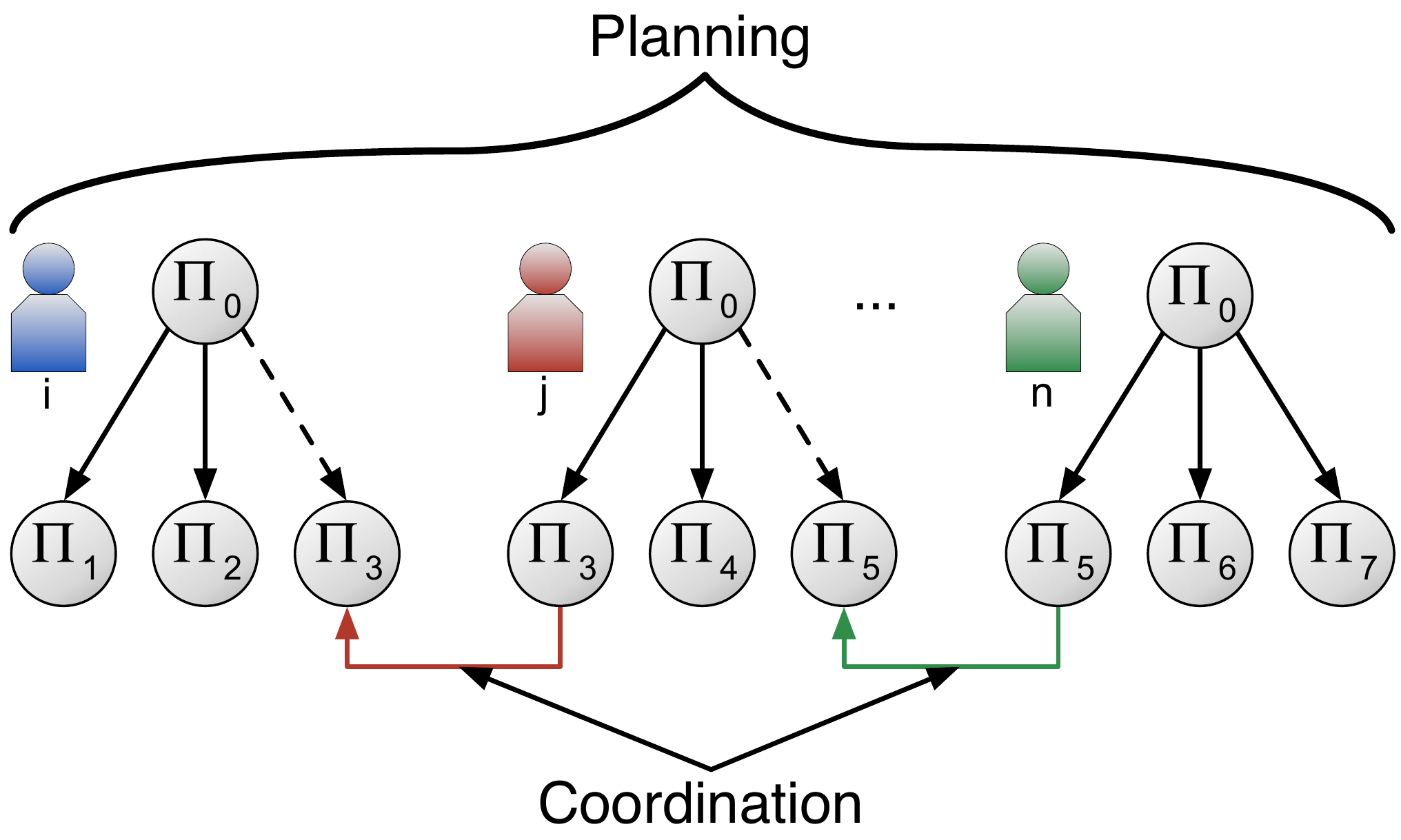}
\caption{Multi-agent search in interleaved planning and coordination}
\label{FigInterleaved}
\end{figure}

\subsubsection{Interleaved Planning and Coordination}
\label{interleaved}

A broad range of MAP techniques \emph{interleave} the planning and coordination activities. This coordination strategy is particularly appropriate for tasks that feature \cgoals since agents explore the search space jointly to find a solution plan, rather than obtaining local solutions individually. In this context, agents continuously coordinate with each other to communicate their findings, thus effectively intertwining planning and coordination.

Most interleaved solvers, such as \MAFS \cite{Nissim12} and \FMAP \cite{Torreno14}, commonly rely on a \emph{coordinated multi-agent search} scheme, wherein nodes of the search space are contributed by several agents (see Figure \ref{FigInterleaved}). This scheme involves selecting a node for expansion (planning) and exchanging the successor nodes among the agents (coordination). Agents thus jointly explore the search space until a solution is found, alternating between phases of planning and coordination. 

Different forms of coordination are applicable in the interleaved resolution strategy. In \FMAP \cite{Torreno14}, agents share an open list of plans and  jointly select the most promising plan according to global heuristic estimates. Each agent $i$ expands the selected node using its actions $\A^i$, and then, agents evaluate and exchange all the successor plans. In \MAFS \cite{Nissim12}, each agent $i$ keeps an independent open list of states. Agents carry out the search simultaneously: an agent $i$ selects a state $S$ to expand from its open list according to a local heuristic estimate, and synthesizes all the nodes that can be generated through the application of the actions $\A^i$ over $S$. Out of all the successor nodes, agents only share the states that are relevant to other agents.

Interleaving planning and coordination is very suitable for solving complex tasks that involve \cgoals and a high coordination effort. By using this strategy, agents learn the cooperation requirements of other participants during the construction of the plan and can immediately address them. Hence, the interleaved scheme allows agents to efficiently address \cgoals.

The main drawback of this coordination strategy is the high communication cost in a distributed MAP setting because alternating planning and coordination usually entails exchanging a high number of messages in order to continuously coordinate agents.

\subsection{Communication mechanisms}
\label{comm}

Communication among agents plays a central role in MAP solvers that conceptually define multiple planners (see Table \ref{conceptual_schemes}), since planning agents must coordinate their activities in order to accomplish the task goals. As shown in Figure \ref{FigDistribution}, different agent communication mechanisms can be applied, depending on the computational process followed by the MAP solver.

\paragraph{Internal communication} Solvers that draw upon a centralized implementation resort to \emph{internal} or simulated multi-agent communication. For example, the centralized solver \MAPR \cite{Borrajo13} distributes the task goals, $\G$, among the planning agents, and agents solve their local tasks sequentially. Once a local plan $\Pi^i$ for an agent $i \in AG$ is computed, the information of $\Pi^i$ is used as an input for the next agent in the sequence, $j$, thus simulating a message passing between agents $i$ and $j$. This type of simple and simulated communication system is all that is required in centralized solvers that run all planning agents in a single machine. 

\paragraph{External communication} As displayed in Figure \ref{FigDistribution} (right), distributed MAP solvers draw upon \emph{external} communication mechanisms by which different processes (agents), potentially allocated on different machines, exchange messages in order to interact with each other. External communication can be easily enabled by linking the different agents via network sockets or a messaging broker. For example, agents in \MAPLAN \cite{maplan-codmap15} exchange data over the TCP/IP protocol when the solver is executed in a distributed manner. A common alternative to implement external communication in MAS implies using a message passing protocol compliant with the IEEE FIPA standards \cite{FIPA02protocol}, which are intended to promote the interoperation of heterogeneous agents. The \MAG MAS platform \cite{Such12} used by \MHFMAP \cite{Torreno15} facilitates the implementation of agents with FIPA-compliant messaging capabilities.

The use of external communication mechanisms allows distributed solvers to run the planning agents in decentralized machines and to coordinate their activities by exchanging messages through the network. The flexibility provided by external communication mechanisms comes at the cost of performance degradation. The results of the 2015 \CODMAP competition \cite{Komenda16codmap} show that centralized solvers like \ADP \cite{Crosby13} outperform solvers executed in a distributed setting, such as \PSM \cite{psm-codmap15}. Likewise, an analysis performed in \cite{Torreno14} reveals that communication among agents is the most time-consuming activity of the distributed approach \FMAP, thus compromising the overall scalability of this solver. Nevertheless, the participants in the distributed track of the \CODMAP exhibited a competitive performance, which proves that the development of fully-distributed MAP solvers is worth the overhead caused by external communication infrastructures.

\subsection{Heuristic Search}
\label{heuristic}

Ever since the introduction of \HSP~\cite{Long00}, the use of heuristic functions that guide the search by estimating the quality of the nodes of the search tree has proven to be one of the most robust and reliable problem-solving strategies in single-agent planning. Over the years, many solvers based on heuristic search, such as \FF~\cite{Hoffmann01} or \LAMA~\cite{Richter10}, have consistently dominated the International Planning Competitions\footnote{http://www.icaps-conference.org/index.php/Main/Competitions}. 

Since most MAP solvers stem from single-agent planning techniques, heuristic search is one of the most common approaches in the MAP literature. In general, in solvers that synthesize a plan for multiple executors, the single planning agent (see Table \ref{conceptual_schemes}) has complete access to the MAP task $\T$, and so it can compute heuristics that leverage the \emph{global} information of $\T$, in a way that is very similar to that of single-agent planners. However, in solvers that feature multiple planning agents, i.e., $|\AG| > 1$, each agent $i$ is only aware of the information defined in $\T^i$ and no agent has access to the complete task $\T$. Under a scheme of planning by multiple agents, one can distinguish between \emph{local} and \emph{global} heuristics.

In local heuristics, an agent $i$ estimates the cost of the task goals, $\G$, using only the information in $\T^i$. The simplicity of local heuristics, which do not require any interactions among agents, contrasts with the low accuracy of the estimates they yield due to the limited task view of the agents. Consider, for instance, the example task $\T_1$ presented in section \ref{characterization}: agent $ta1$ does not have sufficient information to compute an accurate estimate of the cost to reach the goals of $\T_1$ since $\T^{ta1}_1$ does not include the configuration of the area $ga2$. In general, if $\T^i$ is a limited view of $\T$, local heuristics will not yield informative estimates of the cost of reaching $\G$. 

In contrast, a global heuristic in MAP is the application of a heuristic function ``carried out by several agents which have a different knowledge of the task and, possibly, privacy requirements'' \cite{Torreno15}. The development of global heuristics for multi-agent scenarios must account for additional features that make heuristic evaluation an arduous task \cite{Nissim12}: 

\begin{itemize}
\item Solvers based on distributed computation require robust communication protocols for agents to calculate estimates for the overall task.
\item For MAP approaches that preserve agents' privacy, the communication protocol must ensure that estimates are computed without disclosing sensitive private data. 
\end{itemize}

The application of local or global heuristics is also determined by the characteristics of the plan synthesis scheme of the MAP solver. Particularly, in an \emph{unthreaded} planning and coordination scheme, agents synthesize their local plans through the application of local heuristic functions.  For instance, in the sequential plan synthesis scheme of \MAPR \cite{Borrajo13}, agents apply locally $h_{FF}$ \cite{Hoffmann01} and $h_{Land}$ \cite{Richter10} when solving their allocated goals.

Local heuristic search has also been applied by some \emph{interleaved} MAP solvers.  Agents in \MAFS and \MADA \cite{Nissim14} generate and evaluate search states locally. An agent $i$ shares a state $S$ and local estimate $h^i(S)$ only if $S$ is relevant to other planning agents. Upon reception of $S$, agent $j$ performs its local evaluation of $S$, $h^j(S)$. Then, depending on the characteristics of the heuristic, the final estimate of $S$ by agent $j$ will be either $h^i(S)$, $h^j(S)$, or a combination of both.  In \cite{Nissim12}, authors test \MADA with two different optimal heuristics, \lmcut \cite{Helmert09} and \mands \cite{Helmert07}, both locally applied by each agent. Despite heuristics being applied only locally, \MADA is proven to be cost-optimal. 

\vspace{0,2cm}

Unlike unthreaded solvers, the interleaved planning and coordination strategy makes it possible to accommodate global heuristic functions. In this case, agents apply some heuristic function on their tasks $\T^i$ and then they exchange their local estimates to come up with an estimate for the global task $\T$. 

\GPPP \cite{Maliah14} introduces a distributed version of a privacy-preserving \emph{landmark} extraction algorithm for MAP. A landmark is a proposition that must be satisfied in every solution plan of a MAP task \cite{Hoffman04}. The quality of a plan in \GPPP is computed as the sum of the local estimates of the agents in $\AG$. \GPPP outperforms \MAFS thanks to the accurate estimates provided by this landmark-based heuristic. \MHFMAP \cite{Torreno15}, the latest version of \FMAP, introduces a multi-heuristic alternation mechanism based on Fast Downward (\FD) \cite{Helmert06}. Agents alternate two global heuristics when expanding a node in their tasks: $h_{DTG}$, which draws upon the information of the \emph{Domain Transition Graphs} \cite{Helmert04} associated with the state variables of the task, and the landmark-based heuristic $h_{Land}$, which only evaluates the \emph{preferred successors} \cite{Torreno15}. Agents jointly build the DTGs and the landmarks graph of the task and each of them stores its own version of the graphs according to its knowledge of the MAP task.

Some recent work in the literature focuses on the adaptation of well-known single-agent heuristic functions to compute global MAP estimators. The authors of \cite{Stolba14} adapt the single-agent heuristic $h_{FF}$ \cite{Hoffmann01} by means of a compact structure, the \emph{exploration queue}, that optimizes the number of messages exchanged among agents. This multi-agent version of $h_{FF}$, however, is not as accurate as the single-agent one. The work in \cite{Stolba15} introduces a global MAP version of the admissible heuristic function \lmcut that is proven to obtain estimates of the same quality as the single-agent \lmcut. This multi-agent \lmcut yields better estimates at the cost of a larger computational cost. 

In conclusion, heuristic search in MAP, and most notably, the development of global heuristic functions in a distributed context, constitutes one of the main challenges of the MAP research community. The aforementioned approaches prove the potential of the development and combination of global heuristics towards scaling up the performance of MAP solvers. 

\subsection{Privacy}
\label{privacy}

The preservation of agents' sensitive information, or \emph{privacy}, is one of the basic aspects that must be enforced in MAP. The importance of privacy is illustrated in the task $\T_1$ of section \ref{characterization}, which includes two different agents, $ta1$ and $ta2$, both representing a transport agency. Although both agents are meant to cooperate for solving this task, it is unlikely that they are willing to reveal sensitive information to a potential competitor.

Privacy in MAP has been mostly neglected and under-represented in the literature. Some paradigms like Hierarchical Task Network (HTN) planning apply an form of implicit privacy when an agent delegates subgoals to another agent, which solves them by concealing the resolution details from the requester agent. This makes HTN a very well-suited approach for practical applications like composition of web services \cite{Sirin04}. However, formal treatment of privacy is even more scarce. One of the first attempts to come up with a formal privacy model in MAP is found in \cite{vanderKrogt07b}, where authors quantify privacy in terms of the Shannon's information theory \cite{Shannon48}. More precisely, authors establish a notion of uncertainty with respect to plans and provide a measure of privacy loss in terms of the data uncovered by the agents along the planning process. Unfortunately, this measure is not general enough to capture details such as heuristic computation. Nevertheless, quantification of privacy is an important issue in MAP, as it is in distributed constraint satisfaction problems \cite{Faltings08}. A more recent work, also based on Shannon's information theory \cite{Stolba16b}, quantifies privacy leakage for \MASTRIPSF according to the reduction of the number of possible transition systems caused by the revealed information. In this work, the main sources of privacy leakage are identified, but not experimentally evaluated. 

\begin{center}
\begin{table}[h]
\bgroup
\def\arraystretch{1.2}
\centering
\tbl{Categorization of privacy properties in MAP}{
{\footnotesize
\begin{tabular}{| c || l |}
   \hline   
	Privacy criterion & Categories \\ \hline \hline
  \multirow{2}{*}{Modelling of private information} & Imposed privacy \cite{Brafman08} \\ \cline{2-2}
   & Induced privacy \cite{Torreno14}\\ \hline \hline
   \multirow{2}{*}{Information sharing} & \MASTRIPSF \cite{Brafman08}\\ \cline{2-2}
   & Subset privacy \cite{Bonisoli14} \\ \hline \hline
   \multirow{4}{*}{Practical guarantees} & No privacy \cite{Decker92} \\ \cline{2-2}
   & Weak privacy \cite{Borrajo13} \\ \cline{2-2}
   & Object cardinality privacy \cite{Shani16} \\ \cline{2-2}
   & Strong privacy \cite{Brafman15} \\ \hline 
\end{tabular}}
}
\egroup
\label{tab:privacy}
\end{table}
\end{center}

The next subsections analyze the privacy models adopted by the MAP solvers in the literature according to three different criteria (see summary in Table \ref{tab:privacy}): the \emph{modelling} of private information, the \emph{information sharing} schemes, and the privacy \emph{practical privacy guarantees} offered by the MAP solver.

\subsubsection{Modelling of Private Information}

This feature is closely related to whether the language used to specify the MAP task enables explicit modelling of privacy or not. 

Early approaches to MAP, such as \MASTRIPSF \cite{Brafman08}, manage a notion of \emph{induced privacy}. Since the \MASTRIPSL language does not explicitly model private information, the agents' private data are inferred from the task structure. Given an agent $i \in \AG$ and a piece of information $p^i \in \T^i$, $p^i$ is defined as private if $\forall_{j \in \AG | j \neq i} p^i \not \in \T^j$; that is, if $p^i$ is known to $i$ and ignored by the rest of agents in $\T$.

\FMAP \cite{Torreno14} introduces a more general \emph{imposed privacy} scheme, explicitly describing the private and shareable information in the task description. \MAPDDL~\cite{Kovacs12}, the language used in the \CODMAP competition~\cite{Komenda16codmap}, follows this imposed privacy scheme and allows the designer to model the private elements of the agents' tasks. 

In general, both induced and imposed privacy schemes are commonly applied by current MAP solvers. The induced privacy scheme enables the solver to automatically identify the naturally private elements of a MAP task. The imposed privacy scheme, by contrast, offers a higher control and flexibility to model privacy, which is a helpful tool in contexts where agents wish to occlude sensitive data that would be shared otherwise. 

\subsubsection{Information Sharing}

Privacy-preserving algorithms vary accordingly to the number of agents that share a particular piece of information. In general, we can identify two information sharing models, namely \MASTRIPSF and \emph{subset privacy}.

\paragraph{\MASTRIPSF} The \MASTRIPSF information sharing model \cite{Brafman08} defines as public the data that are shared among all the agents in $\AG$, so that a piece of information is either known to all the participants, or only to a single agent. More precisely, a proposition $p \in \Prop$ is defined as \emph{internal} or \emph{private} to an agent $i \in \AG$ if  $p$ is only used and affected by the actions of $\A^i$. However, if the proposition $p$ is also in the preconditions and/or effects of some action $\alpha \in \A^j$, where $j \in \AG$ and $j\neq i$, then $p$ is publicly accessible to all the agents in $\AG$. 

An action $\alpha \in \A^i$ that contains only public preconditions and effects is said to be public and it is known to all the participants in the task. In case $\alpha$ includes both public and private preconditions and effects, agents share instead $\alpha_p$, the \emph{public projection} of $\alpha$, an abstraction that contains only the public elements of $\alpha$.

This simple dichotomic privacy model of \MASTRIPSF does not allow for specifying MAP tasks that require some information to be shared only by a subset of the planning agents in $\AG$.

\paragraph{Subset privacy} Subset privacy is introduced in \cite{Bonisoli14} and generalizes the \MASTRIPSF scheme by establishing pairwise privacy. This model defines a piece of information as private to a single agent, publicly accessible to all the agents in $\AG$ or known to a \emph{subset of agents}. This approach is useful in applications where agents wish to conceal some information from certain agents.

For instance, agent $ta2$ in task $\T_1$ of section \ref{characterization} notifies the factory agent $ft$ whenever the proposition $(pos\;t2\;ft)$ is reached. This proposition indicates that the truck $t2$ is placed at the factory $ft$, a location that is known to both $ta2$ and $ft$. Under the \MASTRIPSF model, agent $ta1$ would be notified that truck $t2$ is at the factory $ft$, an information that $ta2$ may want to conceal. However, the subset privacy model allows $ta2$ to hide $(pos\;t2\;ft)$ from $ta1$ by defining it as private between $ta2$ and $ft$. 

Hence, subset privacy is a more flexible information sharing model compared to the more conservative and limited approach of \MASTRIPSF, which enables representing more complex and realistic situations concerning information sharing.

\subsubsection{Privacy Practical Guarantees}
\label{practical_privacy}

Recent studies devoted to a formal treatment of practical privacy guidelines in MAP \cite{Nissim14,Shani16} conclude that some privacy schemes allow agents to infer private information from other agents through the transmitted data. According to these studies, it is possible to establish a four-level taxonomy to classify the practical privacy level of MAP solvers. The four levels of the taxonomy, from the least to the most secure one, are: \emph{no privacy}, \emph{weak privacy},  \emph{object cardinality privacy} and \emph{strong privacy}. 

\paragraph{No privacy} Privacy has been mostly neglected in MAP but has been extensively treated within the MAS community \cite{Such12}. The 2015 \CODMAP competition introduced a more expressive definition of privacy than \MASTRIPSF and this was a boost for many planners to model private data in the task descriptions. Nevertheless, we can cite a large number of planners that completely disregard the issue of privacy among agents such as early approaches like \GPGP \cite{Decker92} or more recent approaches like \USATPLAN \cite{Dimopoulos12}, \AP~\cite{Jezequel12} or \DPGM \cite{Pellier10}.

\paragraph{Weak privacy} A MAP system is weakly privacy-preserving if agents do not explicitly communicate their private information to other agents at execution time \cite{Brafman15}. This is accomplished by either \emph{obfuscating} (encrypting) or \emph{occluding} the private information they communicate to other agents in order to only reveal the public projection of their actions. In a weak privacy setting, agents may infer private data of other agents through the information exchanged during the plan synthesis. 

\emph{Obfuscating} the private elements of a MAP task is an appropriate mechanism when agents wish to conceal the meaning of propositions and actions. In obfuscation, the proposition names are encrypted but the number and unique identity of preconditions and effects of actions are retained, so agents are able to reconstruct the complete isomorphic image of their tasks. In \MAPR \cite{Borrajo13}, \PMR \cite{Luis14} and \CMAP \cite{mapr-cmap-codmap15}, when an agent communicates a plan, it encrypts the private information in order to preserve its sensitive information. Agents in \MAFS  \cite{Nissim14}, \MADLA \cite{madla-codmap15}, \MAPLAN \cite{maplan-codmap15} and \GPPP \cite{Maliah14} encrypt the private data of the relevant states that they exchange during the plan synthesis. 

Other weak privacy-preserving solvers in the literature \emph{occlude} the agents' private information rather than sharing obfuscated data. Agents in \MHFMAP \cite{Torreno15} only exchange the public projection of the actions of their partial-order plans, thus occluding private information like preconditions, effects, links or orderings.  

\paragraph{Object cardinality privacy} Recently, the \DPP planner \cite{Shani16} introduced a new level of privacy named object cardinality privacy. A MAP algorithm preserves object cardinality privacy if, given an agent $i$ and a type $t$, the cardinality of $i$'s private objects of type $t$ cannot be inferred by other agents from the information they receive \cite{Shani16}. In other words, this level of privacy strongly preserves the number of objects of a given type $t$ of an agent $i$, thus representing a middle ground between the weak and strong privacy settings. 
	
Hiding the cardinality of private objects is motivated by real-world scenarios. Consider, for example, the logistics task $\T_1$ of section \ref{characterization}. One can assume that the transport agencies that take part in the MAP task, $ta1$ and $ta2$, know that packages are delivered using trucks. However, it is likely that each agent would like to hide the number of trucks it possesses or the number of transport routes it uses.

\paragraph{Strong privacy} A MAP algorithm is said to strongly preserve privacy if none of the agents in $\AG$ is able to infer a private element of an agent's task from the public information it obtains during planning. In order to guarantee strong privacy, it is necessary to consider several factors, such as the nature of the communication channel (synchronous, asynchronous, lossy) or the computational power of the agents. 

\SMAFS \cite{Brafman15} is a theoretical proposal to strong privacy that builds upon the \MAFS \cite{Nissim14} model. In \SMAFS, two states that only differ in their private elements are not communicated to other agents in order to prevent them from deducing information through the non-private or public part of the states. \SMAFS is proved to guarantee strong privacy for a sub-class of tasks based on the well-known \emph{logistics} domain. 

\vspace{0,2cm}

In summary, weak privacy is easily achievable through obfuscation of private data, but provides little security. On the other hand, the proposal of \SMAFS lays the theoretical foundations to strong privacy in MAP but the complexity analysis and the practical implementation issues of this approach have not been studied yet. Additionally, object cardinality privacy accounts for a middle ground between weak and strong privacy. In general, the vast majority of MAP methods are classified under the no privacy or weak privacy levels: the former approaches to MAP do not consider privacy at all, while most of the recent proposals, which claim to be privacy preserving, resort in most cases to \emph{obfuscation} to conceal private information.

\begin{center}
\begin{sidewaystable*}
\begin{centering}
\tbl{Summary of the state-of-the-art MAP solvers and their features. For unthreaded solvers, the plan synthesis schemes are listed in form of pairs "agent coordination'' \& "local planning technique".}{
{\scalebox{0.8}
{\scriptsize
\bgroup
\def\arraystretch{1.3}
\begin{tabular}{|l||l|l|l|l|l||l|l|}
\hline
\multirow{2}{*}{{\footnotesize MAP Solver}} & {\footnotesize Coordination} & {\footnotesize Computational} & \multirow{2}{*}{{\footnotesize Plan synthesis scheme}} & \multirow{2}{*}{{\footnotesize Heuristic}} & \multirow{2}{*}{{\footnotesize Privacy}} & \multicolumn{2}{|c|}{\CODMAP} \tabularnewline
\cline{7-8}
& {\footnotesize strategy} & {\footnotesize process} & & & & Cent. track & Dist. track\tabularnewline
\hline
\hline
{\ADP \cite{Crosby13}} & {UT} & {C}  & {Automated task agentization \& heuristic forward search (\FD)} & {L} & {N} & 1st/5th & -  \tabularnewline
\hline
\multirow{2}{*}{\MALAPKT \cite{map-lapkt-codmap15}} & \multirow{2}{*}{UT} & \multirow{2}{*}{C} & {Task mapping into single-agent task \&} & \multirow{2}{*}{G}  & \multirow{2}{*}{W} & \multirow{2}{*}{2nd/3rd/6th} & \multirow{2}{*}{-} \tabularnewline 
 & & & {heuristic forward search (\LAPKT)}  & & & & \tabularnewline
\hline
\multirow{2}{*}{\MARC \cite{marc-codmap15}} & \multirow{2}{*}{UT} & \multirow{2}{*}{C} & {Task mapping into transformer agent task \&  planning via } & \multirow{2}{*}{G} & \multirow{2}{*}{W} & \multirow{2}{*}{4th} & \multirow{2}{*}{-} \tabularnewline 
 & & & {\FD or \IBACOP $\rightarrow$ solution plan translation into original MAP task}  & & & & \tabularnewline
\hline
\multirow{2}{*}{\CMAP \cite{mapr-cmap-codmap15}} & \multirow{2}{*}{UT} & \multirow{2}{*}{C} & {Pre-planning goal allocation $\rightarrow$ task mapping into single-agent task $\rightarrow$} & \multirow{2}{*}{G}  & \multirow{2}{*}{W} & \multirow{2}{*}{7th/8th} & \multirow{2}{*}{-}  \tabularnewline 
 & & & {solution plan parallelization \& heuristic forward search (\LAMA)}  & & & & \tabularnewline
\hline
{\MAPLAN \cite{maplan-codmap15}} & {IL} & {D} & {Multi-agent heuristic forward search} & {G/L} & {W} & 9th/18th/19th & 2nd/5th/6th \tabularnewline
\hline
{\GPPP \cite{Maliah16}} & {IL} & {C} & {Multi-agent heuristic forward search (relaxed, subgoals)} & {G} & {W} & 10th/11th & - \tabularnewline
\hline	
\multirow{2}{*}{\PSM \cite{Tozicka15}} & \multirow{2}{*}{UT} & \multirow{2}{*}{D} & {Intersection of Finite Automata \&} & \multirow{2}{*}{G} & \multirow{2}{*}{W} & \multirow{2}{*}{12th/16th} & \multirow{2}{*}{1st/4th} \tabularnewline 
 & & & {heuristic forward search (\LAMA) $\rightarrow$ Finite Automata}  & & & & \tabularnewline
\hline
{\MADLA \cite{Stolba14}} & {IL} & {C} & {Multi-agent multi-heuristic state-based search} & {G/L} & {W} & 13th & -  \tabularnewline
\hline
\multirow{2}{*}{\PMR \cite{Luis14}} & \multirow{2}{*}{UT} & \multirow{2}{*}{C} & {Pre-planning goal allocation $\rightarrow$ Plan merging $\rightarrow$ plan repair $\rightarrow$} & \multirow{2}{*}{L} & \multirow{2}{*}{W} & \multirow{2}{*}{14th} & \multirow{2}{*}{-} \tabularnewline 
 & & & {solution plan parallelization \& heuristic forward search (\LAMA)}  & & & & \tabularnewline
\hline
\multirow{2}{*}{\MAPR \cite{Borrajo13}} & \multirow{2}{*}{UT} & \multirow{2}{*}{C} & {Pre-planning goal allocation $\rightarrow$ iterative response planning $\rightarrow$} & \multirow{2}{*}{L}  & \multirow{2}{*}{W} & \multirow{2}{*}{15th} & \multirow{2}{*}{-}  \tabularnewline 
 & & & {solution plan parallelization  \& heuristic forward search (\LAMA)}  & & & & \tabularnewline
\hline
\MHFMAP \cite{Torreno15} & {IL} & {D} & {Multi-agent A{*} multi-heuristic search via forward POP} & {G} & {W} & 17th & 3rd  \tabularnewline 
\hline
\hline
\multirow{2}{*}{\DPP \cite{Shani16}} & \multirow{2}{*}{UT} & \multirow{2}{*}{C} & {Synthesis of high-level plan over DP projection (\FD) \&} & \multirow{2}{*}{L} & \multirow{2}{*}{OC} & \multirow{2}{*}{-} & \multirow{2}{*}{-}  \tabularnewline
 & & & heuristic forward search (\FF) & & & & \tabularnewline
\hline
{\FMAP \cite{Torreno14}} & {IL} & {D} & {Multi-agent A{*} heuristic search via forward POP} & {G} & {W} & - & -  \tabularnewline 
\hline
\MAFS \cite{Nissim12} & {IL} & {D} & Multi-agent heuristic forward search & L & W  & - & - \tabularnewline
\hline
{\MADA \cite{Nissim12}} & {IL} & {D} & {Multi-agent A{*} heuristic forward search} & L & W & - & - \tabularnewline
\hline
\MAPPOP \cite{Torreno12KAIS} & {IL} & {D} &  {Multi-agent A{*} heuristic search via backward POP} & G & W & - & -  \tabularnewline 
\hline
{\PF \cite{Brafman08}} & {UT} & {C} & {Post-planning coordination via DisCSP \& heuristic forward search (\FF)} & {--} & {N} & - & - \tabularnewline
\hline
{\USATPLAN \cite{Dimopoulos12}} & {UT} & {C} & {Pre-planning goal allocation $\rightarrow$ iterative response planning \& SAT} & {--} & {N} & - & - \tabularnewline
\hline
\multirow{2}{*}{\TFPOP \cite{Kvarnstrom11}} & \multirow{2}{*}{UT} & \multirow{2}{*}{C} & {Forward-chaining partial-order planning \&} & {--} & \multirow{2}{*}{N} & \multirow{2}{*}{-} & \multirow{2}{*}{-}  \tabularnewline
 & & & {synthesis of agent-specific thread of actions}  & & & & \tabularnewline
\hline
{\DP \cite{Fabre10}} & {UT} & {C} & {Message passing algorithm \& Finite Automata} & {--} & {N} & - & - \tabularnewline
\hline
{\DPGM \cite{Pellier10}} & {UT} & {C} & {Iterative response planning \& GraphPlan + CSP plan extraction} & {--} & {N} & - & - \tabularnewline
\hline
\AP \cite{Jezequel12} & {UT} & {C} & {Asynchronous communication mechanism \& A{*} heuristic forward search} & {G} & {N} & - & -  \tabularnewline
\hline
{\SMAFS \cite{Brafman15}} & {IL} & {C} & {Multi-agent heuristic forward search} & {L} & {S} & - & - \tabularnewline
\hline
\end{tabular}
\egroup
}}}
\par\end{centering}{\footnotesize \par}

\begin{tabnote}
\tabnoteentry{}{Computational process: C - centralized solver, D - distributed solver}
\tabnoteentry{}{Coordination strategy: UT - unthreaded, IL - interleaved}
\tabnoteentry{}{Privacy: N - no privacy, W - weak privacy, OC - object cardinality privacy, S - strong privacy}
\tabnoteentry{}{Heuristic: L - local, G - global}
\tabnoteentry{}{\CODMAP: coverage classification, Cent. track - centralized track, Dist. track - distributed track, a '-' indicates that the solver did not participate in a track.}

\end{tabnote}

\label{tab:features}
\end{sidewaystable*}

\par\end{center}

\section{Distributed and Multi-Agent Planning Systems Taxonomy}
\label{taxonomy}

As discussed in section \ref{introduction}, MAP is a long-running research field that has been covered in several articles \cite{Desjardins99a,deWeerdt09,MeneguzziS15}. This section reviews the large number of domain-independent cooperative MAP solvers that have been proposed since the introduction of the \MASTRIPSF model \cite{Brafman08}. This large body of research was recently crystallized in the 2015 \CODMAP competition \cite{Komenda16codmap}, the first attempt to directly compare MAP solvers through a benchmark encoded using the standardized \MAPDDL language.

The cooperative solvers analyzed in this section cover a wide range of different plan synthesis schemes. As discussed in section \ref{aspects}, one can identify several aspects that determine the features of MAP solvers; namely, \emph{agent distribution}, \emph{computational process}, \emph{plan synthesis scheme}, \emph{communication mechanism}, \emph{heuristic search} and \emph{privacy preservation}. This section presents an in-depth taxonomy that classifies solvers according to their main features and analyzes their similarities and differences (see Table \ref{tab:features}).

This section also aims to critically analyze and compare the strengths and weaknesses of the planners regarding their applicability and experimental performance. Given that a comprehensive comparison of MAP solvers was issued as a result of the 2015 \CODMAP competition, Table \ref{tab:features} arranges solvers according to their positions in the coverage ranking (number of problems solved) of this competition. The approaches included in this taxonomy are organized according to their plan synthesis scheme, an aspect that ultimately determines the types of MAP tasks they can solve. Section \ref{taxonomy_unthreaded} discusses the planners that follow an unthreaded planning and coordination scheme, while section \ref{taxonomy_interleaved} reviews interleaved approaches to MAP.

\subsection{Unthreaded Planning and Coordination MAP Solvers}
\label{taxonomy_unthreaded}

The main characteristic of unthreaded planners is that planning and coordination are not intertwined but handled as two separate and independent activities. Unthreaded solvers are labelled as \emph{UT} in the column \emph{Coordination strategy} of Table \ref{tab:features}. They typically apply local single-agent planning and a combinatorial optimization or satisfiability algorithm to coordinate the local plans.

\subsubsection*{\PF, 2008 (implemented in 2010)}

\PF~\cite{Nissim10} is the first MAP solver that builds upon the \MASTRIPSF model. It is an early representative of the unthreaded strategy that inspired the development of many subsequent MAP solvers, which are presented in the next paragraphs. \PF generates a local plan for each agent in a centralized fashion by means of the \FF planner \cite{Hoffmann01}, and coordinates the local plans through a distributed Constraint Satisfaction Problem (DisCSP) solver to come up with a global solution plan. More precisely, \PF distributes the MAP task among the agents and identifies the coordination points of the task as the actions whose application affects other agents. The DisCSP is then used to find consistent coordination points between the local plans. If the DisCSP solver finds a solution, the plan for the MAP task is directly built from the local plans since the DisCSP solution guarantees compatibility among the underlying local plans.  

The authors of \cite{Nissim10} empirically evaluate \PF over a set of tasks based on the well-known \emph{rovers}, \emph{satellite} and \emph{logistics} domains. The results show that a large number of coordination points among agents derived from the number of public actions limits the scalability and effectiveness of \PF. Later, the \MAPPOP solver outperformed \PF in both execution time and coverage \cite{Torreno12ECAI}.

\subsubsection*{\DPGM, 2010 (implemented in 2013)}

\DPGM \cite{Pellier10} makes also use of CSP techniques to coordinate the agents' local plans. Unlike \PF, the CSP solver in \DPGM is explicitly distributed across agents and it is used to extract the local plans from a set of distributed planning graphs. Under the \emph{iterative response planning} strategy introduced by \DPGM, the solving process is started by one agent, which proposes a local plan along with a set of coordination constraints. The subsequent agent uses its CSP to extract a local plan compatible with the prior agent's plan and constraints. If an agent is not able to generate a compliant plan, \DPGM backtracks to the previous agent, which puts forward an alternative plan with different coordination constraints. 

\subsubsection*{\USATPLAN, 2010}

\USATPLAN \cite{Dimopoulos12} is a MAP solver that extends the satisfiability-based planner \SATPLAN \cite{Kautz06} to a multi-agent context. \USATPLAN performs an \emph{a priori} distribution of the MAP task goals, $\G$, among the agents in $\AG$. Similarly to \DPGM, agents follow an iterative response planning strategy, where each participant takes the previous agent's solution as an input and extends it to solve its assigned goals via \SATPLAN. This way, agents progressively generate a solution.

\USATPLAN is unable to attain tasks that include \cgoals because it assumes that each agent can solve its assigned goals by itself. \USATPLAN is experimentally validated on several multi-agent tasks of the \emph{logistics}, \emph{storage} and \emph{TPP} domains. Although these tasks feature only two planning agents, the authors claim that \USATPLAN is capable of solving tasks with a higher number agents.

\subsubsection*{\MAPR, \PMR, \CMAP, 2013-2015}

Multi-Agent Planning by Reuse (\MAPR) \cite{Borrajo13} allocates the goals $\G$ of the task among the agents before planning through a relaxed reachability analysis. The private information of the local plans is encrypted, thus preserving weak privacy by obfuscating the agents' local tasks. \MAPR also follows an iterative response plan synthesis scheme, wherein an agent takes as input the result of the prior agent's solution plan and runs the \LAMA planner \cite{Richter10} to obtain an extended solution plan that attains its allocated agent's goals.  The plan of the last agent is a solution plan for the MAP task, which is parallelized to ensure that execution agents perform as many actions in parallel as possible. \MAPR is limited to tasks that do not feature specialized agents or \cgoals. This limitation is a consequence of the assumption that each agent is able to solve its allocated goals by itself, which renders \MAPR incomplete.

Plan Merging by Reuse (\PMR) \cite{Luis14,pmr-codmap15} draws upon the goal allocation and obfuscation privacy mechanisms of \MAPR.  Unlike \MAPR, agents carry out the planning stage simultaneously instead of sequentially and each agent generates local plans for its assigned goals. In the post-planning plan merging strategy of \PMR, the resulting local plans are concatenated, yielding a sequential global solution. If the result of the merging process is not a valid solution plan, local plans are merged through a repair procedure. If a merged solution is not found, the task is solved via a single-agent planner. 

Although \CMAP \cite{mapr-cmap-codmap15} follows the same goal allocation and obfuscation strategy of \MAPR and \PMR, the plan synthesis scheme of \CMAP transforms the encrypted local tasks into a single-agent task ($|\AG| = 1$), which is then solved through the planner \LAMA. \CMAP was the best-performing approach of this family of MAP planners in the 2015 \CODMAP competition as shown in Table \ref{tab:features}. \CMAP ranked 7th in the centralized track and exhibited a solid performance over the 12 domains of the \CODMAP benchmark (approximately 90\% coverage). \PMR and \MAPR  ranked 14th and 15th, with roughly 60\% coverage. The plan synthesis scheme of \MAPR affects its performance in domains that feature \cgoals, such as \emph{depots} or \emph{woodworking}, while \PMR offers a more stable performance over the benchmark.

\subsubsection*{\MALAPKT, 2015}

\MALAPKT \cite{map-lapkt-codmap15} conceives a MAP task as a problem that can be transformed and solved by a single-agent planner using the appropriate encoding. More precisely, \MALAPKT compiles the MAP task into a task that features one planning agent ($|\AG|=1$) and solves with the tools provided in the repository \LAPKT \cite{lapkt}. The authors of \cite{map-lapkt-codmap15} try three different variations of best-first and depth-first search that result in algorithms with different theoretical properties and performance. The task translation performed by \MALAPKT offers weak privacy preservation guarantees.

As shown in Table \ref{tab:features}, two of the three versions of \MALAPKT that participated in the \CODMAP ranked 2nd and 3rd in the centralized track. The coverage of \MALAPKT and of \CMAP is roughly to 90\% of the benchmark problems, an indication that shows the efficiency of the scheme that compiles a MAP task into a single-agent task.

\subsubsection*{\MARC, 2015}

The Multi-Agent Planner for Required Cooperation (\MARC) \cite{marc-codmap15} is a centralized MAP solver based on the theory of required cooperation \cite{Zhang14}. \MARC analyzes the agent distribution of the MAP task and comes up with a different arrangement of planning agents. Particularly, \MARC compiles the original task into a task with a set of \emph{transformer agents}, each one being an ensemble of various agents; i.e., $|\AG_{MARC}| < |\AG|$. A transformer agent comprises the representation of various agents of the original MAP task including all their actions. The current implementation of \MARC compiles all the agents in $\AG$ into a single transformer agent ($|\AG_{MARC}| = 1$). Then, a solution plan is computed via \FD \cite{Helmert06} or the portfolio planner \IBACOP \cite{Cenamor14}, and the resulting plan is subsequently translated into a solution for the original MAP task. \MARC preserves weak privacy since private elements of the MAP task are occluded in the transformer agent task.

Regarding experimental performance, \MARC ranks at the 4th position of the centralized \CODMAP with 90\% coverage, thus being one of the best-performing MAP approaches. The experimental results also reveal the efficiency of this multi-to-one agent transformation.

\subsubsection*{\ADP, 2013}

The Agent Decomposition-based Planner (\ADP) \cite{Crosby13} is a factored planning solver that exploits the inherently multi-agent structure (agentization) of some \STRIPS-style planning tasks and comes up with a MAP task where $|\AG| > 1$. \ADP applies a state-based centralized planning procedure to solve the MAP task. In each iteration, \ADP determines a set of subgoals that are achievable from the current state by one of the agents. A search process, guided through the well-known $h_{FF}$ heuristic, is then applied to find a plan that achieves these subgoals, thus resulting in a new state. This mechanism iterates successively until a solution is found.

Experimentally, \ADP outperforms several state-of-the-art classical planners (e.g., \LAMA) and is the top-ranked solver at the centralized track of the \CODMAP, outperforming other approaches that compile the MAP task into a single-agent planning task, such as \MALAPKT or \CMAP.

\subsubsection*{\DP, 2010}

\DP \cite{Fabre10} is a factored planning approach that exploits independence within a planning task. Unlike other factored methods \cite{Kelareva07}, \DP does not set any bound on the number of actions or coordination points of local plans. In \DP, a component or abstraction of the global task is represented as a finite automaton, which recognizes the regular language formed by the local valid plan of the component. This way, all local plans are manipulated at once and a generic distributed optimization technique enables to limit the number of compatible local plans. With this unbounded representation, all valid plans can be computed in one run but stronger conditions are required to guarantee polynomial runtime.  \DP is the first optimal MAP solver in the literature (note that \PF is optimal with respect to the number of coordination points, but local planning is carried out through a suboptimal planner).

\DP was experimentally tested in a factored version of the \emph{pipesworld} domain. However, the solver is unable to solve even the smallest instances of this domain in a reasonable time. The authors claim that the reason is that \DP scales roughly as $n^3$, where $n$ is the number of components of the global task. For obvious reasons, \DP has not been empirically compared against other MAP solvers in the literature.

\subsubsection*{\AP, 2012 (not implemented)}

In the line of factored planning, \AP \cite{Jezequel12} is a multi-agent A* search that finds a path for the goal in each local component of a task and ensures that the component actions that must be jointly performed are compatible. \AP runs in parallel a modified version of the A* algorithm in each component, and the local search processes are guided towards finding local plans that are compatible with each other. Each local A* finds a plan as a path search in a graph and informs its neighbors of the common actions that may lead to a solution. Particularly, each agent searches its local graph or component while considering the constraints and costs of the rest of agents, received through an asynchronous communication mechanism. The authors of \cite{Jezequel12} do not validate \AP experimentally; however, the soundness, completeness and optimality properties of \AP are formally proven.

\subsubsection*{\PSM, 2014}

\PSM \cite{Tozicka15,psm-codmap15} is a recent distributed MAP solver that follows \DP's compact representation of local agents plans into Finite Automata, called Planning State Machines (PSMs). This planner defines two basic operations: obtaining a public projection of a PSM and merging two different PSMs. These operations are applied to build a public PSM consisting of merged public parts of individual PSMs. The plan synthesis scheme gradually expands the agents' local PSMs by means of new local plans. A solution for the MAP task is found once the public PSM is not empty. \PSM weakly preserves privacy as it obfuscates states of the PSMs in some situations.

\PSM applies an efficient handling of communication among agents, which grants this solver a remarkable experimental performance in both the centralized and distributed setting. In the centralized \CODMAP track, \PSM ranks 12th (solving 70\% of the tasks), and it is the top performer at the distributed track of the competition.

\subsubsection*{\DPP, 2016}

The DP-Projection Planner (\DPP) \cite{Shani16}, is a centralized \MASTRIPSF solver that uses the Dependency-Preserving (DP) projection, a novel and accurate public projection of the MAP task information with object cardinality privacy guarantees. The single planning agent of \DPP uses the \FD planner to synthesize a high-level plan which is then extended with the agents' private actions via the \FF planner, thus resulting in a multi-agent solution plan.

The authors of \cite{Shani16} provide a comprehensive experimental evaluation of \DPP through the complete benchmark of the 2015 \CODMAP competition. The results show that \DPP outperforms most of the top contenders of the competition (namely, \GPPP, \MAPR, \PMR, \MAPLAN and \PSM). All in all, \DPP can be considered the current top \MASTRIPSF-based solver, as well as one of the best-performing MAP approaches to date.

\subsubsection*{\TFPOP, 2011}

\TFPOP \cite{Kvarnstrom11} is a hybrid approach  that combines the flexibility of partial-order planning and the performance of forward-chaining search. Unlike most \MASTRIPSF-based solvers, \TFPOP supports temporal reasoning with durative actions. \TFPOP is a centralized approach that synthesizes a solution for multiple executors. It computes \emph{threaded partial-order plans}; i.e., non-linear plans that keep a thread of sequentially-ordered actions per agent, since authors assume that an execution agent performs its actions sequentially. 

\TFPOP is tested in a reduced set of domains, which include the well-known \emph{satellite} and \emph{zenotravel} domains, as well as a UAV delivery domain. The objective of this experimentation is to compare \TFPOP against several partial-order planners. \TFPOP is not compared to any MAP solver in this taxonomy.

\subsection{Interleaved Planning and Coordination MAP Solvers}
\label{taxonomy_interleaved}

Under the interleaved scheme, labelled as \emph{IL} in the column \emph{Coordination strategy} of Table \ref{tab:features}, agents jointly explore the search space intertwining their planning and coordination activities. The development of interleaved MAP solvers heavily relies on the design of robust communication protocols to coordinate agents during planning.

\subsubsection*{\MAPPOP, \FMAP, \MHFMAP, 2010-2015}

In this family of MAP solvers, agents apply a distributed exploration of the plan space. Agents locally compute plans through an embedded partial-order planning (POP) component and they build a joint search tree by following an A* search scheme guided by global heuristic functions.

\MAPPOP \cite{Torreno12ECAI,Torreno12KAIS} performs an incomplete search based on a classical backward POP algorithm and POP heuristics. \FMAP \cite{Torreno14} introduces a sound and complete plan synthesis scheme that uses a forward-chaining POP \cite{Benton12} guided through the $h_{DTG}$ heuristic. \MHFMAP \cite{fmap-codmap15} applies a multi-heuristic search approach that alternates $h_{DTG}$ and $h_{Land}$, building a Landmark Graph (LG) to estimate the number of pending landmarks of the partial-order plans. The three planners guarantee weak privacy since private information is occluded throughout the planning process and heuristic evaluation. The $h_{Land}$ estimator uses some form of obfuscation during the construction of the LG.

Regarding experimental results, \FMAP is proven to outperform \MAPPOP and \MAPR in terms of coverage over 10 MAP domains, most of which are included in the \CODMAP benchmark. Results in \cite{Torreno15} indicate that \MHFMAP obtains better coverage than both \FMAP and \GPPP. Interestingly, this planner exhibits a much worse performance in the \CODMAP (see Table \ref{tab:features}), ranking 17th with only 42\% coverage. This is due to the lose of accuracy of the $h_{DTG}$ heuristic when the internal state-variable representation of the tasks in \MHFMAP is transformed to a propositional representation to be tested in the \CODMAP benchmark, thus compromising the performance of the solver \cite{fmap-codmap15}.

\subsubsection*{\MADLA, 2013}

The Multiagent Distributed and Local Asynchronous Planner (\MADLA) \cite{Stolba15} is a centralized solver that runs one thread per agent on a single machine and combines two versions of the $h_{FF}$ heuristic, a projected (local) variant ($h_L$) and a distributed (global) variant ($h_D$) in a multi-heuristic state-space search. The main novelty of \MADLA is that the agent which is computing $h_D$, which requires contributions of the other agents for calculating the global heuristic estimator, is run asynchronously and so it can continue the search using $h_L$ while waiting for responses from other agents that are computing parts of $h_D$. \MADLA evaluates as many states as possible using the global heuristic $h_D$, which is more informative than $h_L$. This way, \MADLA can use a computationally hard global heuristic without blocking the local planning process of the agents, thus improving the performance of the system.

Experimentally, \MADLA ranks 13th in the centralized \CODMAP, reporting 66\% coverage. It outperforms most of the distributed MAP solvers of the competition, but it is not able to solve the most complex tasks of the \CODMAP domains, thus not reaching the figures of the top performers such as \ADP, \MALAPKT or \MARC.

\subsubsection*{\MAFS, \MADA, 2012-2014}

\MAFS \cite{Nissim14} is an updated version of \PF that implements a distributed algorithm wherein agents apply a heuristic state-based search (see section \ref{heuristic}). In \cite{Nissim12}, authors present \MADA, a cost-optimal variation of \MAFS. In this case, each agent expands the state that minimizes $f=g+h$, where $h$ is estimated through an admissible heuristic. Particularly, authors tested the landmark heuristic \emph{LM-Cut} \cite{Helmert09} and the abstraction heuristic \emph{Merge\&Shrink}~\cite{Helmert07}. \MADA is the first distributed and interleaved solver based on \MASTRIPSF.

\MAFS is compared against \MAPPOP and \PF over the \emph{logistics}, \emph{rovers} and \emph{satellite} domains, notably outperforming both solvers in terms of coverage and execution time \cite{Nissim14}. On the other hand, the authors of \cite{Nissim14} only compare \MADA against single-agent optimal solvers.

\subsubsection*{\SMAFS, 2015 (not implemented)}

\SMAFS \cite{Brafman15} is an extension of \MAFS towards secure MAP, and it is currently the only solver that offers \emph{strong privacy} guarantees (see section \ref{practical_privacy}). Currently, \SMAFS is a theoretical work that has not been yet implemented nor experimentally evaluated.

\subsubsection*{\GPPP, 2014}

The Greedy Privacy-Preserving Planner (\GPPP) \cite{Maliah14,Maliah16} builds upon \MAFS and improves its performance via a global landmark-based heuristic function. \GPPP applies a \emph{global planning} stage and then a \emph{local planning} stage. In the former, agents agree on a joint coordination scheme by solving a relaxed MAP task that only contains public actions (thereby preserving privacy) and obtaining a skeleton plan. In the \emph{local planning} stage, agents compute private plans to achieve the preconditions of the actions in the skeleton plan. Since coordination is done over a relaxed MAP task, the individual plans of the agents may not succeed at solving the actions' preconditions. In this case, the global planning stage is executed again to generate a different coordination scheme, until a solution is found. In \GPPP, agents weakly preserve privacy by obfuscating the private information of the shared states through private state identifiers.

\GPPP provides a notable experimental performance, ranking 10th in the centralized \CODMAP track. \GPPP reaches 83\% coverage and is only surpassed by the different versions of \ADP, \MARC, \MALAPKT, \CMAP and \MAPLAN, which proves the accuracy of its landmark based heuristic and the overall efficiency of its plan synthesis scheme.

\subsubsection*{\MAPLAN, 2015}

\MAPLAN \cite{maplan-codmap15} is a heuristic \MAFS-based solver that adapts several concepts from \MADA and \MADLA. \MAPLAN is a distributed and flexible approach that implements a collection of state-space search methods, such as best first or A{*}, as well as several local and global heuristic functions ($h_{FF}$, \emph{LM-Cut}, potential heuristics and others), which allows the solver to be run under different configurations. \MAPLAN can be executed in a single-machine, using local communication, or in a distributed fashion, where each agent is in a different machine and communication among agents is implemented through network message passing. Regarding privacy, \MAPLAN applies a form of obfuscation, replacing private facts in search states by unique local identifiers, which grants weak privacy. 

\MAPLAN exhibits a very solid performance in the centralized and distributed tracks of the 2015 \CODMAP competition, ranking 9th and 2nd, respectively. In the centralized track, \MAPLAN obtains 83\% coverage, outperforming \GPPP and reaching similar figures than the top-performing centralized solvers.

\section{Conclusions}
\label{ongoing}

The purpose of this article is to comprehensively survey the state of the art in cooperative MAP, offering an in-detail overview of this rapidly evolving research field, which has experienced multiple key advances over the last decade. These contributions crystallized in the 2015 \CODMAP competition, where MAP solvers were compared through an exhaustive benchmark testing encoded with \MAPDDL, the first standard modelling language for MAP tasks.

In this paper, the topic of MAP was studied from a twofold perspective: from the representational structure of a MAP task and from the problem-solving standpoint. We formally defined a MAP task following the well-known MA-STRIPS model and provided several examples which illustrate the features that distinguish MAP tasks from the more compact single-agent planning tasks. We also presented the modelling of these illustrative tasks with  \MAPDDL. 

MAP is a broad field that allows for a wide variety of problem-solving approaches. For this reason, we identified and thoroughly analyzed the main aspects that characterize a solver, from the architectural design to the practical features of a MAP tool. Among others, these aspects include the computational process of the solvers and the plan synthesis schemes that stem from the particular combination of planning and coordination applied by MAP tools, as well as other key features, such as the communication mechanisms used by the agents to interact with each other and the privacy guarantees offered by the existing solvers.

Finally, we compiled and classified the existing MAP techniques according to the aforementioned criteria. The taxonomy of MAP techniques presented in this survey prioritizes recent domain-independent techniques in the literature. Particularly, we focused on the approaches that took part in the 2015 \CODMAP competition, comparing their performance, strengths and weaknesses. The classification aims to provide the reader with a clear and comprehensive overview of the existing cooperative MAP solvers.

\vspace{0,2cm}

The body of work presented in this survey constitutes a solid foundation for the ongoing and future scientific development of the MAP field. Following, we summarize several research trends that have recently captured the attention of the community.

\paragraph{Theoretical properties} The aim of the earlier cooperative MAP solvers was to contribute with a satisficing approach capable of solving a relatively small number of problems in a reasonable time but without providing any formal properties \cite{Nissim10,Borrajo13}. The current maturity of the cooperative MAP field has witnessed the introduction of some models that focus on granting specific theoretical properties, such as completeness \cite{Torreno14}, optimality \cite{Nissim14} or stronger privacy preservation guarantees \cite{Brafman15,Shani16}.

\paragraph{Privacy} The state of the art in MAP shows a growing effort in analyzing and formalizing privacy in MAP solvers. Nowadays, various approaches to model private information and to define information sharing can be found in the literature, which reveals that privacy is progressively becoming a key topic in MAP. However, the particular implementation of a MAP solver may jeopardize privacy, if it is possible for an agent to infer private information from the received public data. Aside from the four-level classification exposed in section \ref{practical_privacy}, other recent approaches attempt to theoretically quantify the privacy guarantees of a MAP solver \cite{Stolba16b}. In the same line, the authors of \cite{Tozicka17} analyze the implications and limits of strong privacy and present a novel \PSM-based planner that offers strong privacy guarantees.

On the other hand, one can also find work that proposes a smart use of privacy to increase the performance of MAP solvers,  like \DPP \cite{Shani16}, which calculates an accurate public projection of the MAP task in order to obtain a robust high-level plan that is then completed with private actions.  This scheme minimizes the communication requirements, resulting in a more efficient search. In \cite{Maliah17}, authors introduce a novel weak privacy-preserving variant of \MAFS which ensures that two agents that do not share any private variable never communicate with each other, significantly reducing the number of exchanged messages. In general, the study of privacy in MAP is gaining much attention and more and more sophisticated approaches have been recently proposed. 

\paragraph{MAP with self-interesed agents} The mainstream in MAP with self-interested planning agents is handling situations which involve interactive decision making with possibly conflicting interests. Game theory, the study of mathematical models of conflict and cooperation between rational self-interested agents, arises naturally as a paradigm to address human conflict and cooperation within a competitive situation. Game-theoretic MAP is an active and interesting research field that reflects many real-world situations, and thus, it has a broad variety of applications, among which we can highlight congestion games \cite{JonssonR11}, cost-optimal planning \cite{NissimB13}, conflict resolution in the search of a joint plan \cite{JordanO15} or auction systems \cite{RobuNPS11}. 

From a practical perspective, game-theoretic MAP has been successfully applied to ridesharing problems on timetabled transport services \cite{HrncirRJ15}. In general, strategic approaches to MAP are very appropriate to model smart city applications like traffic congestion prevention: vehicles can be accurately modelled as rational self-interested agents that want to reach their destinations as soon as possible, but they are also willing to deviate from their optimal routes in order to avoid traffic congestion issues that would affect all the involved agents.

\paragraph{Practical applications} MAP is being used in a great variety of applications, like in product assembly problems in industry (e.g., car assembly). Agents plan the manufacturing path of the product through the assembly line, which is composed of a number of interconnected resources that can perform different operations. \EXPLANTECH, for instance, is a consolidated framework for agent-based production planning, manufacturing, simulation and supply chain management \cite{Pechoucek07}.

MAP has also been used to control the flow of electricity in the Smart Grid \cite{Reddy11}. The agents' actions are individually rational and contribute to desirable global goals such as promoting the use of renewable energy, encouraging energy efficiency and enabling distributed fault tolerance. Another interesting application of MAP is the automated creation of workflows in biological pathways like the Multi-Agent System for Genomic Annotation (\BIOMAS) \cite{Decker02}. This system uses \DECAF, a toolkit that provides standard services to integrate agent capabilities, and incorporates the \GPGP framework \cite{Lesser04} to coordinate multi-agent tasks.

In decentralized control problems, MAP is applied in coordination of space rovers and helicopter flights, multi-access broadcast channels, and sensor network management, among others \cite{Seuken08}. MAP combined with argumentation techniques to handle belief changes about the context has been used in applications of ambient intelligence in the field of healthcare \cite{Pajares13}.

\vspace{0,2cm}

Aside from the aforementioned trends, there is still a broad variety of unexplored research topics in MAP. The solvers presented in this survey do not support tasks with advanced requirements. Particularly, handling temporal MAP tasks is an unresolved matter that should be addressed in the years to come. This problem will involve both the design of MAP solvers that explicitly support temporal reasoning and the extension of \MAPDDL to incorporate the appropriate syntax to model tasks with temporal constraints.

Cooperative MAP, as exposed in this paper, puts the focus on offline tasks, without paying much attention to the problematic of plan execution. Online planning carried out by several agents poses a series of challenges derived from the integration of planning and execution and the need to respond in complex, real-time environments. Real-time cooperative MAP is about planning and simultaneous execution by several cooperative agents in a changing environment. This interesting and exciting research line is very relevant in applications that involve, for example, soccer robots. 

Additionally, the body of work presented in this survey  does not consider agents with individual preferences. Preference-based MAP is an unstudied field that can be interpreted as a middle ground between cooperative and self-interested MAP, since it involves a set of rational agents that work together towards a common goal while having their own preferences concerning the properties of the solution plan.

All in all, we believe that the steps taken over the last years towards the standardization of MAP tasks and tools, such as the 2015 \CODMAP competition or the introduction of \MAPDDL, will decisively contribute to foster a rapid expansion of this field in a wide variety of research directions.





\bibliographystyle{ACM-Reference-Format-Journals}
\bibliography{references}


\begin{thebibliography}{00}


\ifx \showCODEN    \undefined \def \showCODEN     #1{\unskip}     \fi
\ifx \showDOI      \undefined \def \showDOI       #1{{\tt DOI:}\penalty0{#1}\ }
  \fi
\ifx \showISBNx    \undefined \def \showISBNx     #1{\unskip}     \fi
\ifx \showISBNxiii \undefined \def \showISBNxiii  #1{\unskip}     \fi
\ifx \showISSN     \undefined \def \showISSN      #1{\unskip}     \fi
\ifx \showLCCN     \undefined \def \showLCCN      #1{\unskip}     \fi
\ifx \shownote     \undefined \def \shownote      #1{#1}          \fi
\ifx \showarticletitle \undefined \def \showarticletitle #1{#1}   \fi
\ifx \showURL      \undefined \def \showURL       #1{#1}          \fi

\bibitem[\protect\citeauthoryear{Amir and Engelhardt}{Amir and
  Engelhardt}{2003}]%
        {Amir03}
{Eyal Amir} {and} {Barbara Engelhardt}. 2003.
\newblock \showarticletitle{Factored planning}. In {\em Proceedings of the 18th
  International Joint Conference on Artificial Intelligence (IJCAI)}, Vol.~3.
  929--935.
\newblock


\bibitem[\protect\citeauthoryear{Benton, Coles, and Coles}{Benton
  et~al\mbox{.}}{2012}]%
        {Benton12}
{J. Benton}, {Amanda~J. Coles}, {and} {Andrew~I. Coles}. 2012.
\newblock \showarticletitle{Temporal Planning with Preferences and
  Time-Dependent Continuous Costs}. In {\em Proceedings of the 22nd
  International Conference on Automated Planning and Scheduling (ICAPS)}.
  2--10.
\newblock


\bibitem[\protect\citeauthoryear{Bonisoli, Gerevini, Saetti, and
  Serina}{Bonisoli et~al\mbox{.}}{2014}]%
        {Bonisoli14}
{Andrea Bonisoli}, {Alfonso~E. Gerevini}, {Alessandro Saetti}, {and} {Ivan
  Serina}. 2014.
\newblock \showarticletitle{A Privacy-preserving Model for the Multi-agent
  Propositional Planning Problem}. In {\em Proceedings of the 21st European
  Conference on Artificial Intelligence (ECAI)}. 973--974.
\newblock


\bibitem[\protect\citeauthoryear{Borrajo}{Borrajo}{2013}]%
        {Borrajo13}
{Daniel Borrajo}. 2013.
\newblock \showarticletitle{Multi-Agent Planning by Plan Reuse}. In {\em
  Proceedings of the 12th International Conference on Autonomous Agents and
  Multi-agent Systems (AAMAS)}. 1141--1142.
\newblock


\bibitem[\protect\citeauthoryear{Borrajo and Fern\'{a}ndez}{Borrajo and
  Fern\'{a}ndez}{2015}]%
        {mapr-cmap-codmap15}
{Daniel Borrajo} {and} {Susana Fern\'{a}ndez}. 2015.
\newblock \showarticletitle{{MAPR and CMAP}}. In {\em Proceedings of the
  Competition of Distributed and Multi-Agent Planners (CoDMAP-15)}. 1--3.
\newblock


\bibitem[\protect\citeauthoryear{Boutilier and Brafman}{Boutilier and
  Brafman}{2001}]%
        {BoutilierB01}
{Craig Boutilier} {and} {Ronen~I. Brafman}. 2001.
\newblock \showarticletitle{Partial-Order Planning with Concurrent Interacting
  Actions}.
\newblock {\em Journal of Artificial Intelligence Research\/}  {14} (2001),
  105--136.
\newblock


\bibitem[\protect\citeauthoryear{Brafman}{Brafman}{2015}]%
        {Brafman15}
{Ronen~I. Brafman}. 2015.
\newblock \showarticletitle{A Privacy Preserving Algorithm for Multi-Agent
  Planning and Search}. In {\em Proceedings of the 24th International Joint
  Conference on Artificial Intelligence (IJCAI)}. 1530--1536.
\newblock


\bibitem[\protect\citeauthoryear{Brafman and Domshlak}{Brafman and
  Domshlak}{2006}]%
        {Brafman06}
{Ronen~I. Brafman} {and} {Carmel Domshlak}. 2006.
\newblock \showarticletitle{Factored Planning: How, When, and When Not}. In
  {\em Proceedings of the 21st National Conference on Artificial Intelligence
  and the 18th Innovative Applications of Artificial Intelligence Conference}.
  809--814.
\newblock


\bibitem[\protect\citeauthoryear{Brafman and Domshlak}{Brafman and
  Domshlak}{2008}]%
        {Brafman08}
{Ronen~I. Brafman} {and} {Carmel Domshlak}. 2008.
\newblock \showarticletitle{From One to Many: Planning for Loosely Coupled
  Multi-Agent Systems}. In {\em Proceedings of the 18th International
  Conference on Automated Planning and Scheduling (ICAPS)}. 28--35.
\newblock


\bibitem[\protect\citeauthoryear{Cenamor, de~la Rosa, and
  Fern{\'a}ndez}{Cenamor et~al\mbox{.}}{2014}]%
        {Cenamor14}
{Isabel Cenamor}, {Tom{\'a}s de~la Rosa}, {and} {Fernando Fern{\'a}ndez}. 2014.
\newblock \showarticletitle{{IBACOP} and {IBACOP2} planner}. In {\em
  Proceedings of the International Planning Competition (IPC)}.
\newblock


\bibitem[\protect\citeauthoryear{Clement and Durfee}{Clement and
  Durfee}{1999}]%
        {Clement99}
{Bradley~J. Clement} {and} {Edmund~H. Durfee}. 1999.
\newblock \showarticletitle{Top-down Search for Coordinating the Hierarchical
  Plans of Multiple Agents}. In {\em Proceedings of the 3rd Annual Conference
  on Autonomous Agents} {\em (AGENTS '99)}. ACM, New York, NY, USA, 252--259.
\newblock
\showISBNx{1-58113-066-X}


\bibitem[\protect\citeauthoryear{Corkill}{Corkill}{1979}]%
        {Corkill79}
{Daniel~D. Corkill}. 1979.
\newblock \showarticletitle{Hierarchical Planning in a Distributed
  Environment}. In {\em Proceedings of the Sixth International Joint Conference
  on Artificial Intelligence, {IJCAI} 79, Tokyo, Japan, August 20-23, 1979, 2
  Volumes}. 168--175.
\newblock


\bibitem[\protect\citeauthoryear{Cox and Durfee}{Cox and Durfee}{2004}]%
        {Cox04}
{Jeffrey~S. Cox} {and} {Edmund~H. Durfee}. 2004.
\newblock \showarticletitle{Efficient Mechanisms for Multiagent Plan Merging}.
  In {\em Proceedings of the 3rd Conference on Autonomous Agents and Multiagent
  Systems (AAMAS)}. 1342--1343.
\newblock


\bibitem[\protect\citeauthoryear{Cox and Durfee}{Cox and Durfee}{2009}]%
        {Cox09}
{Jeffrey~S. Cox} {and} {Edmund~H. Durfee}. 2009.
\newblock \showarticletitle{Efficient and distributable methods for solving the
  multiagent plan coordination problem}.
\newblock {\em Multiagent and Grid Systems\/} {5}, 4 (2009), 373--408.
\newblock


\bibitem[\protect\citeauthoryear{Crosby, Jonsson, and Rovatsos}{Crosby
  et~al\mbox{.}}{2014}]%
        {Crosby14}
{Matthew Crosby}, {Anders Jonsson}, {and} {Michael Rovatsos}. 2014.
\newblock \showarticletitle{A Single-Agent Approach to Multiagent Planning}. In
  {\em Proceedings of the 21st European Conference on Artificial Intelligence
  (ECAI)}. 237--242.
\newblock


\bibitem[\protect\citeauthoryear{Crosby, Rovatsos, and Petrick}{Crosby
  et~al\mbox{.}}{2013}]%
        {Crosby13}
{Matthew Crosby}, {Michael Rovatsos}, {and} {Ronald P.~A. Petrick}. 2013.
\newblock \showarticletitle{Automated Agent Decomposition for Classical
  Planning}. In {\em Proceedings of the 23rd International Conference on
  Automated Planning and Scheduling (ICAPS)}. 46--54.
\newblock


\bibitem[\protect\citeauthoryear{de~Weerdt, Bos, Tonino, and
  Witteveen}{de~Weerdt et~al\mbox{.}}{2003}]%
        {WeerdtBTW03}
{Mathijs de Weerdt}, {Andr{\'{e}} Bos}, {Hans Tonino}, {and} {Cees Witteveen}.
  2003.
\newblock \showarticletitle{A Resource Logic for Multi-Agent Plan Merging}.
\newblock {\em Annals of Mathematics and Artificial Intelligence\/} {37}, 1-2
  (2003), 93--130.
\newblock


\bibitem[\protect\citeauthoryear{de~Weerdt and Clement}{de~Weerdt and
  Clement}{2009}]%
        {deWeerdt09}
{Mathijs de Weerdt} {and} {Bradley~J. Clement}. 2009.
\newblock \showarticletitle{Introduction to planning in multiagent systems}.
\newblock {\em Multiagent and Grid Systems\/} {5}, 4 (2009), 345--355.
\newblock


\bibitem[\protect\citeauthoryear{Decker, Khan, Schmidt, Situ, Makkena, and
  Michaud}{Decker et~al\mbox{.}}{2002}]%
        {Decker02}
{Keith Decker}, {Salim Khan}, {Carl Schmidt}, {Gang Situ}, {Ravi Makkena},
  {and} {Dennis Michaud}. 2002.
\newblock \showarticletitle{BioMAS: a multi-agent system for genomic
  annotation}.
\newblock {\em International Journal of Cooperative Information Systems\/}
  {11}, 3 (2002), 265--292.
\newblock


\bibitem[\protect\citeauthoryear{Decker and Lesser}{Decker and Lesser}{1992}]%
        {Decker92}
{Keith Decker} {and} {Victor~R. Lesser}. 1992.
\newblock \showarticletitle{Generalizing the {P}artial {G}lobal {P}lanning
  Algorithm}.
\newblock {\em International Journal of Cooperative Information Systems\/} {2},
  2 (1992), 319--346.
\newblock


\bibitem[\protect\citeauthoryear{desJardins and Wolverton}{desJardins and
  Wolverton}{1999}]%
        {Desjardins99b}
{Marie desJardins} {and} {Michael Wolverton}. 1999.
\newblock \showarticletitle{Coordinating a Distributed Planning System}.
\newblock {\em {AI} Magazine\/} {20}, 4 (1999), 45--53.
\newblock


\bibitem[\protect\citeauthoryear{desJardins, Durfee, Ortiz, and
  Wolverton}{desJardins et~al\mbox{.}}{1999}]%
        {Desjardins99a}
{Marie~E. desJardins}, {Edmund~H. Durfee}, {Charles~L. Ortiz}, {and}
  {Michael~J. Wolverton}. 1999.
\newblock \showarticletitle{A survey of research in distributed continual
  planning}.
\newblock {\em AI Magazine\/} {20}, 4 (1999), 13--22.
\newblock


\bibitem[\protect\citeauthoryear{Dimopoulos, Hashmi, and Moraitis}{Dimopoulos
  et~al\mbox{.}}{2012}]%
        {Dimopoulos12}
{Yannis Dimopoulos}, {Muhammad~A. Hashmi}, {and} {Pavlos Moraitis}. 2012.
\newblock \showarticletitle{$\mu$-SATPLAN: Multi-agent planning as
  satisfiability}.
\newblock {\em Knowledge-Based Systems\/}  {29} (2012), 54--62.
\newblock


\bibitem[\protect\citeauthoryear{Dix, Mu{\~{n}}oz{-}Avila, Nau, and Zhang}{Dix
  et~al\mbox{.}}{2003}]%
        {DixMNZ03}
{J{\"{u}}rgen Dix}, {H{\'{e}}ctor Mu{\~{n}}oz{-}Avila}, {Dana~S. Nau}, {and}
  {Lingling Zhang}. 2003.
\newblock \showarticletitle{IMPACTing {SHOP:} Putting an {AI} Planner Into a
  Multi-Agent Environment}.
\newblock {\em Annals of Mathematics and Artificial Intelligence\/} {37}, 4
  (2003), 381--407.
\newblock


\bibitem[\protect\citeauthoryear{Durfee}{Durfee}{1999}]%
        {Durfee99}
{Edmund~H. Durfee}. 1999.
\newblock {\em Distributed problem solving and planning}. Vol. In Gerhard Weiss
  editor.
\newblock The MIT Press, San Francisco, CA, 118--149.
\newblock


\bibitem[\protect\citeauthoryear{Durfee and Lesser}{Durfee and Lesser}{1991}]%
        {Durfee91}
{Edmund~H. Durfee} {and} {Victor Lesser}. 1991.
\newblock \showarticletitle{{P}artial {G}lobal {P}lanning: A Coordination
  Framework for Distributed Hypothesis Formation}.
\newblock {\em IEEE Transactions on Systems, Man, and Cybernetics, Special
  Issue on Distributed Sensor Networks\/} {21}, 5 (1991), 1167--1183.
\newblock


\bibitem[\protect\citeauthoryear{Ephrati and Rosenschein}{Ephrati and
  Rosenschein}{1994}]%
        {EphratiR94}
{Eithan Ephrati} {and} {Jeffrey~S. Rosenschein}. 1994.
\newblock \showarticletitle{Divide and Conquer in Multi-Agent Planning}. In
  {\em Proceedings of the 12th National Conference on Artificial Intelligence
  (AAAI)}. 375--380.
\newblock


\bibitem[\protect\citeauthoryear{Ephrati and Rosenschein}{Ephrati and
  Rosenschein}{1997}]%
        {Ephrati97}
{Eithan Ephrati} {and} {Jeffrey~S. Rosenschein}. 1997.
\newblock \showarticletitle{A Heuristic Technique for Multi-Agent Planning}.
\newblock {\em Annals of Mathematics and Artificial Intelligence\/} {20}, 1-4
  (1997), 13--67.
\newblock


\bibitem[\protect\citeauthoryear{Fabre, Jezequel, Haslum, and
  Thi{\'{e}}baux}{Fabre et~al\mbox{.}}{2010}]%
        {Fabre10}
{Eric Fabre}, {Lo{\"{\i}}g Jezequel}, {Patrik Haslum}, {and} {Sylvie
  Thi{\'{e}}baux}. 2010.
\newblock \showarticletitle{Cost-Optimal Factored Planning: Promises and
  Pitfalls}. In {\em Proceedings of the 20th International Conference on
  Automated Planning and Scheduling ({ICAPS})}. 65--72.
\newblock


\bibitem[\protect\citeauthoryear{Faltings, L{\'e}aut{\'e}, and Petcu}{Faltings
  et~al\mbox{.}}{2008}]%
        {Faltings08}
{Boi Faltings}, {Thomas L{\'e}aut{\'e}}, {and} {Adrian Petcu}. 2008.
\newblock \showarticletitle{Privacy guarantees through distributed constraint
  satisfaction}. In {\em Proceedings of the 2008 IEEE/WIC/ACM International
  Conference on Web Intelligence and Intelligent Agent Technology (WI-IAT)},
  Vol.~2. IEEE, 350--358.
\newblock


\bibitem[\protect\citeauthoryear{Fikes and Nilsson}{Fikes and Nilsson}{1971}]%
        {Fikes71}
{Richard Fikes} {and} {Nils~J. Nilsson}. 1971.
\newblock \showarticletitle{{STRIPS}: A new approach to the application of
  theorem proving to problem solving}.
\newblock {\em Artificial Intelligence\/} {2}, 3 (1971), 189--208.
\newblock


\bibitem[\protect\citeauthoryear{Fi{\v{s}}er, {\v{S}}tolba, and
  Komenda}{Fi{\v{s}}er et~al\mbox{.}}{2015}]%
        {maplan-codmap15}
{Daniel Fi{\v{s}}er}, {Michal {\v{S}}tolba}, {and} {Anton\'{\i}n Komenda}.
  2015.
\newblock \showarticletitle{{MAPlan}}. In {\em Proceedings of the Competition
  of Distributed and Multi-Agent Planners (CoDMAP-15)}. 8--10.
\newblock


\bibitem[\protect\citeauthoryear{for Intelligent Physical~Agents}{for
  Intelligent Physical~Agents}{2002}]%
        {FIPA02protocol}
{Foundation for Intelligent Physical~Agents}. 2002.
\newblock {FIPA} Interaction Protocol Specification.
  http://www.fipa.org/repository/ips.php3.
\newblock   (2002).
\newblock


\bibitem[\protect\citeauthoryear{Fox and Long}{Fox and Long}{2003}]%
        {Fox03}
{Maria Fox} {and} {Derek Long}. 2003.
\newblock \showarticletitle{{PDDL2.1}: an Extension to {PDDL} for Expressing
  Temporal Planning Domains}.
\newblock {\em Journal of Artificial Intelligence Research\/}  {20} (2003),
  61--124.
\newblock


\bibitem[\protect\citeauthoryear{Ghallab, Howe, Knoblock, McDermott, Ram,
  Veloso, Weld, and Wilkins}{Ghallab et~al\mbox{.}}{1998}]%
        {Ghallab98}
{Malik Ghallab}, {Adele Howe}, {Craig Knoblock}, {Drew McDermott}, {Ashwin
  Ram}, {Manuela~M. Veloso}, {Daniel Weld}, {and} {David Wilkins}. 1998.
\newblock \showarticletitle{{PDDL} - The {P}lanning {D}omain {D}efinition
  {L}anguage}.
\newblock {\em AIPS-98 Planning Committee\/} (1998).
\newblock


\bibitem[\protect\citeauthoryear{Ghallab, Nau, and Traverso}{Ghallab
  et~al\mbox{.}}{2004}]%
        {Ghallab04}
{Malik Ghallab}, {Dana Nau}, {and} {Paolo Traverso}. 2004.
\newblock {\em Automated Planning. Theory and Practice}.
\newblock Morgan Kaufmann.
\newblock


\bibitem[\protect\citeauthoryear{Grosz, Hunsberger, and Kraus}{Grosz
  et~al\mbox{.}}{1999}]%
        {GroszHK99}
{Barbara~J. Grosz}, {Luke Hunsberger}, {and} {Sarit Kraus}. 1999.
\newblock \showarticletitle{Planning and Acting Together}.
\newblock {\em {AI} Magazine\/} {20}, 4 (1999), 23--34.
\newblock


\bibitem[\protect\citeauthoryear{Helmert}{Helmert}{2004}]%
        {Helmert04}
{Malte Helmert}. 2004.
\newblock \showarticletitle{A Planning Heuristic Based on Causal Graph
  Analysis}.
\newblock {\em Proceedings of the 14th International Conference on Automated
  Planning and Scheduling (ICAPS)\/} (2004), 161--170.
\newblock


\bibitem[\protect\citeauthoryear{Helmert}{Helmert}{2006}]%
        {Helmert06}
{Malte Helmert}. 2006.
\newblock \showarticletitle{The {F}ast {D}ownward planning system}.
\newblock {\em Journal of Artificial Intelligence Research\/} {26}, 1 (2006),
  191--246.
\newblock


\bibitem[\protect\citeauthoryear{Helmert and Domshlak}{Helmert and
  Domshlak}{2009}]%
        {Helmert09}
{Malte Helmert} {and} {Carmel Domshlak}. 2009.
\newblock \showarticletitle{Landmarks, Critical Paths and Abstractions: What's
  the Difference Anyway?}. In {\em Proceedings of the 19th International
  Conference on Automated Planning and Scheduling (ICAPS)}. 162--169.
\newblock


\bibitem[\protect\citeauthoryear{Helmert, Haslum, and Hoffmann}{Helmert
  et~al\mbox{.}}{2007}]%
        {Helmert07}
{Malte Helmert}, {Patrik Haslum}, {and} {J{\"o}rg Hoffmann}. 2007.
\newblock \showarticletitle{Flexible Abstraction Heuristics for Optimal
  Sequential Planning}. In {\em Proceedings of the 17th International
  Conference on Automated Planning and Scheduling (ICAPS)}. 176--183.
\newblock


\bibitem[\protect\citeauthoryear{Hoffmann and Nebel}{Hoffmann and
  Nebel}{2001}]%
        {Hoffmann01}
{J{\"o}rg Hoffmann} {and} {Bernhard Nebel}. 2001.
\newblock \showarticletitle{The {FF} Planning System: Fast Planning Generation
  Through Heuristic Search}.
\newblock {\em Journal of Artificial Intelligence Research\/}  {14} (2001),
  253--302.
\newblock


\bibitem[\protect\citeauthoryear{Hoffmann, Porteous, and Sebasti\'a}{Hoffmann
  et~al\mbox{.}}{2004}]%
        {Hoffman04}
{J{\"o}rg Hoffmann}, {Julie Porteous}, {and} {Laura Sebasti\'a}. 2004.
\newblock \showarticletitle{Ordered landmarks in planning}.
\newblock {\em Journal of Artificial Intelligence Research\/}  {22} (2004),
  215--278.
\newblock


\bibitem[\protect\citeauthoryear{Hrnc{\'{\i}}r, Rovatsos, and
  Jakob}{Hrnc{\'{\i}}r et~al\mbox{.}}{2015}]%
        {HrncirRJ15}
{Jan Hrnc{\'{\i}}r}, {Michael Rovatsos}, {and} {Michal Jakob}. 2015.
\newblock \showarticletitle{Ridesharing on Timetabled Transport Services: {A}
  Multiagent Planning Approach}.
\newblock {\em Journal of Intelligent Transportation Systems\/} {19}, 1 (2015),
  89--105.
\newblock


\bibitem[\protect\citeauthoryear{Jezequel and Fabre}{Jezequel and
  Fabre}{2012}]%
        {Jezequel12}
{Lo{\"{\i}}g Jezequel} {and} {Eric Fabre}. 2012.
\newblock \showarticletitle{A{\#}: {A} distributed version of {A*} for factored
  planning}. In {\em Proceedings of the 51th {IEEE} Conference on Decision and
  Control, ({CDC})}. 7377--7382.
\newblock


\bibitem[\protect\citeauthoryear{Jonsson and Rovatsos}{Jonsson and
  Rovatsos}{2011}]%
        {JonssonR11}
{Anders Jonsson} {and} {Michael Rovatsos}. 2011.
\newblock \showarticletitle{Scaling Up Multiagent Planning: A Best-Response
  Approach}. In {\em Proceedings of the 21st International Conference on
  Automated Planning and Scheduling (ICAPS)}. AAAI, 114--121.
\newblock


\bibitem[\protect\citeauthoryear{Jord{\'{a}}n and Onaindia}{Jord{\'{a}}n and
  Onaindia}{2015}]%
        {JordanO15}
{Jaume Jord{\'{a}}n} {and} {Eva Onaindia}. 2015.
\newblock \showarticletitle{Game-Theoretic Approach for Non-Cooperative
  Planning}. In {\em Proceedings of the 29th Conference on Artificial
  Intelligence (AAAI)}. 1357--1363.
\newblock


\bibitem[\protect\citeauthoryear{Kabanza, Shuyun, and Goodwin}{Kabanza
  et~al\mbox{.}}{2004}]%
        {KabanzaSG04}
{Froduald Kabanza}, {Lu Shuyun}, {and} {Scott Goodwin}. 2004.
\newblock \showarticletitle{Distributed Hierarchical Task Planning on a Network
  of Clusters}. In {\em Proceedings of the 16th International Conference on
  Parallel and Distributed Computing and Systems (PDCS)}. 139--140.
\newblock


\bibitem[\protect\citeauthoryear{Kautz}{Kautz}{2006}]%
        {Kautz06}
{Henry~A. Kautz}. 2006.
\newblock \showarticletitle{Deconstructing planning as satisfiability}. In {\em
  Proceedings of the National Conference on Artificial Intelligence}, Vol.~21.
  Menlo Park, CA; Cambridge, MA; London; AAAI Press; MIT Press; 1999, 1524.
\newblock


\bibitem[\protect\citeauthoryear{Kelareva, Buffet, Huang, and
  Thi{\'{e}}baux}{Kelareva et~al\mbox{.}}{2007}]%
        {Kelareva07}
{Elena Kelareva}, {Olivier Buffet}, {Jinbo Huang}, {and} {Sylvie
  Thi{\'{e}}baux}. 2007.
\newblock \showarticletitle{Factored Planning Using Decomposition Trees}. In
  {\em Proceedings of the 20th International Joint Conference on Artificial
  Intelligence (IJCAI)}. 1942--1947.
\newblock


\bibitem[\protect\citeauthoryear{Komenda, Stolba, and Kovacs}{Komenda
  et~al\mbox{.}}{2016}]%
        {Komenda16codmap}
{Anton{\'\i}n Komenda}, {Michal Stolba}, {and} {Daniel~L Kovacs}. 2016.
\newblock \showarticletitle{The International Competition of Distributed and
  Multiagent Planners (CoDMAP)}.
\newblock {\em AI Magazine\/} {37}, 3 (2016), 109--115.
\newblock


\bibitem[\protect\citeauthoryear{Kovacs}{Kovacs}{2012}]%
        {Kovacs12}
{Daniel~L. Kovacs}. 2012.
\newblock \showarticletitle{A Multi-Agent Extension of {PDDL3.1}}. In {\em
  Proceedings of the 3rd Workshop on the International Planning Competition
  (IPC)}. 19--27.
\newblock


\bibitem[\protect\citeauthoryear{Kvarnstr{\"o}m}{Kvarnstr{\"o}m}{2011}]%
        {Kvarnstrom11}
{Jonas Kvarnstr{\"o}m}. 2011.
\newblock \showarticletitle{Planning for Loosely Coupled Agents Using Partial
  Order Forward-Chaining}. In {\em Proceedings of the 21st International
  Conference on Automated Planning and Scheduling (ICAPS)}. AAAI, 138--145.
\newblock


\bibitem[\protect\citeauthoryear{Lesser, Decker, Wagner, Carver, Garvey,
  Horling, Neiman, Podorozhny, Prasad, Raja, Vincent, Xuan, and Zhang}{Lesser
  et~al\mbox{.}}{2004}]%
        {Lesser04}
{Victor Lesser}, {Keith Decker}, {Thomas Wagner}, {Norman Carver}, {Alan
  Garvey}, {Bryan Horling}, {Daniel Neiman}, {Rodion Podorozhny}, {M.~Nagendra
  Prasad}, {Anita Raja}, {Regis Vincent}, {Ping Xuan}, {and} {X.~Q. Zhang}.
  2004.
\newblock \showarticletitle{Evolution of the {GPGP}/{TAEMS} domain-independent
  coordination framework}.
\newblock {\em Autonomous Agents and Multi-Agent Systems\/} {9}, 1-2 (2004),
  87--143.
\newblock


\bibitem[\protect\citeauthoryear{Long, Kautz, Selman, Bonet, Geffner, Koehler,
  Brenner, Hoffmann, Rittinger, Anderson, Weld, Smith, Fox, and Long}{Long
  et~al\mbox{.}}{2000}]%
        {Long00}
{Derek Long}, {Henry Kautz}, {Bart Selman}, {Blai Bonet}, {Hector Geffner},
  {Jana Koehler}, {Michael Brenner}, {Joerg Hoffmann}, {Frank Rittinger},
  {Corin~R. Anderson}, {Daniel~S. Weld}, {David~E. Smith}, {Maria Fox}, {and}
  {Derek Long}. 2000.
\newblock \showarticletitle{The AIPS-98 planning competition}.
\newblock {\em AI magazine\/} {21}, 2 (2000), 13--33.
\newblock


\bibitem[\protect\citeauthoryear{Luis and Borrajo}{Luis and Borrajo}{2014}]%
        {Luis14}
{Nerea Luis} {and} {Daniel Borrajo}. 2014.
\newblock \showarticletitle{Plan Merging by Reuse for Multi-Agent Planning}. In
  {\em Proceedings of the 2nd ICAPS Workshop on Distributed and Multi-Agent
  Planning (DMAP)}. 38--44.
\newblock


\bibitem[\protect\citeauthoryear{Luis and Borrajo}{Luis and Borrajo}{2015}]%
        {pmr-codmap15}
{Nerea Luis} {and} {Daniel Borrajo}. 2015.
\newblock \showarticletitle{{PMR: Plan Merging by Reuse}}. In {\em Proceedings
  of the Competition of Distributed and Multi-Agent Planners (CoDMAP-15)}.
  11--13.
\newblock


\bibitem[\protect\citeauthoryear{Maliah, Brafman, and Shani}{Maliah
  et~al\mbox{.}}{2017}]%
        {Maliah17}
{Shlomi Maliah}, {Ronen~I. Brafman}, {and} {Guy Shani}. 2017.
\newblock \showarticletitle{Increased Privacy with Reduced Communication in
  Multi-Agent Planning}. In {\em Proceedings of the 27th International
  Conference on Automated Planning and Scheduling (ICAPS)}. 209--217.
\newblock


\bibitem[\protect\citeauthoryear{Maliah, Shani, and Stern}{Maliah
  et~al\mbox{.}}{2014}]%
        {Maliah14}
{Shlomi Maliah}, {Guy Shani}, {and} {Roni Stern}. 2014.
\newblock \showarticletitle{Privacy Preserving Landmark Detection}. In {\em
  Proceedings of the 21st European Conference on Artificial Intelligence
  (ECAI)}. 597--602.
\newblock


\bibitem[\protect\citeauthoryear{Maliah, Shani, and Stern}{Maliah
  et~al\mbox{.}}{2016}]%
        {Maliah16}
{Shlomi Maliah}, {Guy Shani}, {and} {Roni Stern}. 2016.
\newblock \showarticletitle{Collaborative privacy preserving multi-agent
  planning}.
\newblock {\em Autonomous Agents and Multi-Agent Systems\/} (2016), 1--38.
\newblock
\showISSN{1573-7454}


\bibitem[\protect\citeauthoryear{Meneguzzi and de~Silva}{Meneguzzi and
  de~Silva}{2015}]%
        {MeneguzziS15}
{Felipe Meneguzzi} {and} {Lavindra de Silva}. 2015.
\newblock \showarticletitle{Planning in {BDI} agents: a survey of the
  integration of planning algorithms and agent reasoning}.
\newblock {\em The Knowledge Engineering Review\/} {30}, 1 (2015), 1--44.
\newblock


\bibitem[\protect\citeauthoryear{Muise, Lipovetzky, and Ramirez}{Muise
  et~al\mbox{.}}{2015}]%
        {map-lapkt-codmap15}
{Christian Muise}, {Nir Lipovetzky}, {and} {Miquel Ramirez}. 2015.
\newblock \showarticletitle{{MAP-LAPKT: Omnipotent Multi-Agent Planning via
  Compilation to Classical Planning}}. In {\em Proceedings of the Competition
  of Distributed and Multi-Agent Planners (CoDMAP-15)}. 14--16.
\newblock


\bibitem[\protect\citeauthoryear{Nau, Au, Ilghami, Kuter, Murdock, Wu, and
  Yaman}{Nau et~al\mbox{.}}{2003}]%
        {NauAIKMWY03}
{Dana~S. Nau}, {Tsz{-}Chiu Au}, {Okhtay Ilghami}, {Ugur Kuter}, {J.~William
  Murdock}, {Dan Wu}, {and} {Fusun Yaman}. 2003.
\newblock \showarticletitle{{SHOP2:} An {HTN} Planning System}.
\newblock {\em Journal of Artificial Intelligence Research\/}  {20} (2003),
  379--404.
\newblock


\bibitem[\protect\citeauthoryear{Nissim and Brafman}{Nissim and
  Brafman}{2012}]%
        {Nissim12}
{Raz Nissim} {and} {Ronen~I. Brafman}. 2012.
\newblock \showarticletitle{Multi-agent {A}* for parallel and distributed
  systems}. In {\em Proceedings of the 11th International Conference on
  Autonomous Agents and Multiagent Systems (AAMAS)}. 1265--1266.
\newblock


\bibitem[\protect\citeauthoryear{Nissim and Brafman}{Nissim and
  Brafman}{2013}]%
        {NissimB13}
{Raz Nissim} {and} {Ronen~I. Brafman}. 2013.
\newblock \showarticletitle{Cost-Optimal Planning by Self-Interested Agents}.
  In {\em Proceedings of the 27th Conference on Artificial Intelligence
  (AAAI)}.
\newblock


\bibitem[\protect\citeauthoryear{Nissim and Brafman}{Nissim and
  Brafman}{2014}]%
        {Nissim14}
{Raz Nissim} {and} {Ronen~I. Brafman}. 2014.
\newblock \showarticletitle{Distributed Heuristic Forward Search for
  Multi-agent Planning}.
\newblock {\em Journal of Artificial Intelligence Research\/}  {51} (2014),
  293--332.
\newblock


\bibitem[\protect\citeauthoryear{Nissim, Brafman, and Domshlak}{Nissim
  et~al\mbox{.}}{2010}]%
        {Nissim10}
{Raz Nissim}, {Ronen~I. Brafman}, {and} {Carmel Domshlak}. 2010.
\newblock \showarticletitle{A general, fully distributed multi-agent planning
  algorithm}. In {\em Proceedings of the 9th International Conference on
  Autonomous Agents and Multiagent Systems (AAMAS)}. 1323--1330.
\newblock


\bibitem[\protect\citeauthoryear{Pajares and Onaindia}{Pajares and
  Onaindia}{2013}]%
        {Pajares13}
{Sergio Pajares} {and} {Eva Onaindia}. 2013.
\newblock \showarticletitle{Context-aware multi-agent planning in intelligent
  environments}.
\newblock {\em Information Sciences\/}  {227} (2013), 22--42.
\newblock


\bibitem[\protect\citeauthoryear{Pechoucek, Reh{\'{a}}k, Charv{\'{a}}t, Vlcek,
  and Kolar}{Pechoucek et~al\mbox{.}}{2007}]%
        {Pechoucek07}
{Michal Pechoucek}, {Martin Reh{\'{a}}k}, {Petr Charv{\'{a}}t},
  {Tom{\'{a}}{\v{s}} Vlcek}, {and} {Michal Kolar}. 2007.
\newblock \showarticletitle{Agent-based approach to mass-oriented production
  planning: case study}.
\newblock {\em {IEEE} Transactions on Systems, Man, and Cybernetics, Part
  {C}\/} {37}, 3 (2007), 386--395.
\newblock


\bibitem[\protect\citeauthoryear{Pellier}{Pellier}{2010}]%
        {Pellier10}
{Damien Pellier}. 2010.
\newblock \showarticletitle{Distributed Planning through Graph Merging}. In
  {\em Proceedings of the 2nd International Conference on Agents and Artificial
  Intelligence (ICAART 2010)}. 128--134.
\newblock


\bibitem[\protect\citeauthoryear{Ramirez, Lipovetzky, and Muise}{Ramirez
  et~al\mbox{.}}{2015}]%
        {lapkt}
{Miquel Ramirez}, {Nir Lipovetzky}, {and} {Christian Muise}. 2015.
\newblock {Lightweight Automated Planning ToolKiT}.
\newblock http://lapkt.org/.   (2015).
\newblock


\bibitem[\protect\citeauthoryear{Reddy and Veloso}{Reddy and Veloso}{2011}]%
        {Reddy11}
{Prashant~P. Reddy} {and} {Manuela~M. Veloso}. 2011.
\newblock \showarticletitle{Strategy learning for autonomous agents in smart
  grid markets}. In {\em Proceedings of the 22nd International Joint Conference
  on Artificial Intelligence (IJCAI)}. 1446--1451.
\newblock


\bibitem[\protect\citeauthoryear{Richter and Westphal}{Richter and
  Westphal}{2010}]%
        {Richter10}
{Silvia Richter} {and} {Matthias Westphal}. 2010.
\newblock \showarticletitle{The {LAMA} planner: Guiding cost-based anytime
  planning with landmarks}.
\newblock {\em Journal of Artificial Intelligence Research\/} {39}, 1 (2010),
  127--177.
\newblock


\bibitem[\protect\citeauthoryear{Robu, Noot, Poutr{\'{e}}, and van
  Schijndel}{Robu et~al\mbox{.}}{2011}]%
        {RobuNPS11}
{Valentin Robu}, {Han Noot}, {Han~La Poutr{\'{e}}}, {and} {Willem{-}Jan van
  Schijndel}. 2011.
\newblock \showarticletitle{A multi-agent platform for auction-based allocation
  of loads in transportation logistics}.
\newblock {\em Expert Systems with Applications\/} {38}, 4 (2011), 3483--3491.
\newblock


\bibitem[\protect\citeauthoryear{Sapena, Onaindia, Garrido, and
  Arang{\'u}}{Sapena et~al\mbox{.}}{2008}]%
        {Sapena08}
{\'Oscar Sapena}, {Eva Onaindia}, {Antonio Garrido}, {and} {Marlene
  Arang{\'u}}. 2008.
\newblock \showarticletitle{A distributed {CSP} approach for collaborative
  planning systems}.
\newblock {\em Engineering Applications of Artificial Intelligence\/} {21}, 5
  (2008), 698--709.
\newblock


\bibitem[\protect\citeauthoryear{Serrano, Such, Bot\'ia, and
  Garc\'ia-Fornes}{Serrano et~al\mbox{.}}{2013}]%
        {Such13}
{Emilio Serrano}, {Jose~M. Such}, {Juan~A. Bot\'ia}, {and} {Ana
  Garc\'ia-Fornes}. 2013.
\newblock \showarticletitle{Strategies for avoiding preference profiling in
  agent-based e-commerce environments}.
\newblock {\em Applied Intelligence\/} (2013), 1--16.
\newblock


\bibitem[\protect\citeauthoryear{Seuken and Zilberstein}{Seuken and
  Zilberstein}{2008}]%
        {Seuken08}
{Sven Seuken} {and} {Shlomo Zilberstein}. 2008.
\newblock \showarticletitle{Formal models and algorithms for decentralized
  decision making under uncertainty}.
\newblock {\em Autonomous Agents and Multi-Agent Systems\/} {17}, 2 (2008),
  190--250.
\newblock


\bibitem[\protect\citeauthoryear{Shani, Maliah, and Stern}{Shani
  et~al\mbox{.}}{2016}]%
        {Shani16}
{Guy Shani}, {Shlomi Maliah}, {and} {Roni Stern}. 2016.
\newblock \showarticletitle{Stronger Privacy Preserving Projections for
  Multi-Agent Planning}. In {\em Proceedings of the 26th International
  Conference on Automated Planning and Scheduling (ICAPS)}. 221--229.
\newblock


\bibitem[\protect\citeauthoryear{Shannon}{Shannon}{1948}]%
        {Shannon48}
{Claude~E. Shannon}. 1948.
\newblock \showarticletitle{A mathematical theory of communication}.
\newblock {\em The Bell System Technical Journal\/} {27}, 3 (1948), 379--423.
\newblock


\bibitem[\protect\citeauthoryear{Sirin, Parsia, Wu, Hendler, and Nau}{Sirin
  et~al\mbox{.}}{2004}]%
        {Sirin04}
{Evren Sirin}, {Bijan Parsia}, {Dan Wu}, {James Hendler}, {and} {Dana Nau}.
  2004.
\newblock \showarticletitle{{HTN} Planning for Web Service Composition using
  {SHOP2}}.
\newblock {\em Journal of Web Semantics\/} {1}, 4 (2004), 377--396.
\newblock


\bibitem[\protect\citeauthoryear{Sreedharan, Zhang, and Kambhampati}{Sreedharan
  et~al\mbox{.}}{2015}]%
        {marc-codmap15}
{Sarath Sreedharan}, {Yu Zhang}, {and} {Subbarao Kambhampati}. 2015.
\newblock \showarticletitle{{A First Multi-agent Planner for Required
  Cooperation (MARC)}}. In {\em Proceedings of the Competition of Distributed
  and Multi-Agent Planners (CoDMAP-15)}. 17--20.
\newblock


\bibitem[\protect\citeauthoryear{{\v{S}}tolba, Fi{\v{s}}er, and
  Komenda}{{\v{S}}tolba et~al\mbox{.}}{2015}]%
        {Stolba15}
{Michal {\v{S}}tolba}, {Daniel Fi{\v{s}}er}, {and} {Anton{\'\i}n Komenda}.
  2015.
\newblock \showarticletitle{Admissible Landmark Heuristic for Multi-Agent
  Planning}. In {\em Proceedings of the 25th International Conference on
  Automated Planning and Scheduling (ICAPS)}. 211--219.
\newblock


\bibitem[\protect\citeauthoryear{{\v{S}}tolba and Komenda}{{\v{S}}tolba and
  Komenda}{2014}]%
        {Stolba14}
{Michal {\v{S}}tolba} {and} {Anton\'in Komenda}. 2014.
\newblock \showarticletitle{Relaxation Heuristics for Multiagent Planning}. In
  {\em Proceedings of the 24th International Conference on Automated Planning
  and Scheduling (ICAPS)}. 298--306.
\newblock


\bibitem[\protect\citeauthoryear{{\v{S}}tolba and Komenda}{{\v{S}}tolba and
  Komenda}{2015}]%
        {madla-codmap15}
{Michal {\v{S}}tolba} {and} {Anton\'{\i}n Komenda}. 2015.
\newblock \showarticletitle{{MADLA: Planning with Distributed and Local
  Search}}. In {\em Proceedings of the Competition of Distributed and
  Multi-Agent Planners (CoDMAP-15)}. 21--24.
\newblock


\bibitem[\protect\citeauthoryear{{\v{S}}tolba, To{\v{z}}i{\v{c}}ka, and
  Komenda}{{\v{S}}tolba et~al\mbox{.}}{2016}]%
        {Stolba16b}
{Michal {\v{S}}tolba}, {Jan To{\v{z}}i{\v{c}}ka}, {and} {Anton{\'\i}n Komenda}.
  2016.
\newblock \showarticletitle{Quantifying Privacy Leakage in Multi-Agent
  Planning}.
\newblock {\em Proceedings of the 4rd ICAPS Workshop on Distributed and
  Multi-Agent Planning (DMAP)\/} (2016), 80--88.
\newblock


\bibitem[\protect\citeauthoryear{Such, Garc{\'\i}a-Fornes, Espinosa, and
  Bellver}{Such et~al\mbox{.}}{2012}]%
        {Such12}
{Jose~M. Such}, {Ana Garc{\'\i}a-Fornes}, {Agust\'in Espinosa}, {and} {Joan
  Bellver}. 2012.
\newblock \showarticletitle{Magentix2: A privacy-enhancing agent platform}.
\newblock {\em Engineering Applications of Artificial Intelligence\/} (2012),
  96--109.
\newblock


\bibitem[\protect\citeauthoryear{Tambe}{Tambe}{1997}]%
        {Tambe97}
{Milind Tambe}. 1997.
\newblock \showarticletitle{Towards Flexible Teamwork}.
\newblock {\em Journal of Artificial Intelligence Research\/}  {7} (1997),
  83--124.
\newblock


\bibitem[\protect\citeauthoryear{Torre\~no, Onaindia, and Sapena}{Torre\~no
  et~al\mbox{.}}{2012}]%
        {Torreno12ECAI}
{Alejandro Torre\~no}, {Eva Onaindia}, {and} {\'Oscar Sapena}. 2012.
\newblock \showarticletitle{An approach to multi-agent planning with incomplete
  information}. In {\em Proceedings of the 20th European Conference on
  Artificial Intelligence (ECAI)}, Vol. 242. IOS Press, 762--767.
\newblock


\bibitem[\protect\citeauthoryear{Torre\~no, Onaindia, and Sapena}{Torre\~no
  et~al\mbox{.}}{2014a}]%
        {Torreno12KAIS}
{Alejandro Torre\~no}, {Eva Onaindia}, {and} {\'Oscar Sapena}. 2014a.
\newblock \showarticletitle{{A} Flexible Coupling Approach to Multi-Agent
  Planning under Incomplete Information}.
\newblock {\em Knowledge and Information Systems\/} {38}, 1 (2014), 141--178.
\newblock


\bibitem[\protect\citeauthoryear{Torre\~no, Onaindia, and Sapena}{Torre\~no
  et~al\mbox{.}}{2014b}]%
        {Torreno14}
{Alejandro Torre\~no}, {Eva Onaindia}, {and} {\'Oscar Sapena}. 2014b.
\newblock \showarticletitle{{FMAP}: Distributed cooperative multi-agent
  planning}.
\newblock {\em Applied Intelligence\/} {41}, 2 (2014), 606--626.
\newblock


\bibitem[\protect\citeauthoryear{Torre\~no, Onaindia, and Sapena}{Torre\~no
  et~al\mbox{.}}{2015}]%
        {Torreno15}
{Alejandro Torre\~no}, {Eva Onaindia}, {and} {\'Oscar Sapena}. 2015.
\newblock \showarticletitle{Global Heuristics for Distributed Cooperative
  Multi-Agent Planning}. In {\em Proceedings of the 25th International
  Conference on Automated Planning and Scheduling (ICAPS)}. 225--233.
\newblock


\bibitem[\protect\citeauthoryear{Torre\~{n}o, Sapena, and Onaindia}{Torre\~{n}o
  et~al\mbox{.}}{2015}]%
        {fmap-codmap15}
{Alejandro Torre\~{n}o}, {\'{O}scar Sapena}, {and} {Eva Onaindia}. 2015.
\newblock \showarticletitle{{MH-FMAP: Alternating Global Heuristics in
  Multi-Agent Planning}}. In {\em Proceedings of the Competition of Distributed
  and Multi-Agent Planners (CoDMAP-15)}. 25--28.
\newblock


\bibitem[\protect\citeauthoryear{To{\v{z}}i{\v{c}}ka, Jakubuv, and
  Komenda}{To{\v{z}}i{\v{c}}ka et~al\mbox{.}}{2015}]%
        {psm-codmap15}
{Jan To{\v{z}}i{\v{c}}ka}, {Jan Jakubuv}, {and} {Anton\'{\i}n Komenda}. 2015.
\newblock \showarticletitle{{PSM-based Planners Description for CoDMAP 2015
  Competition}}. In {\em Proceedings of the Competition of Distributed and
  Multi-Agent Planners (CoDMAP-15)}. 29--32.
\newblock


\bibitem[\protect\citeauthoryear{To{\v{z}}i{\v{c}}ka, Jakubuv, Komenda, and
  P{\v{e}}chou{\v{c}}ek}{To{\v{z}}i{\v{c}}ka et~al\mbox{.}}{2016}]%
        {Tozicka15}
{Jan To{\v{z}}i{\v{c}}ka}, {Jan Jakubuv}, {Anton{\'{\i}}n Komenda}, {and}
  {Michal P{\v{e}}chou{\v{c}}ek}. 2016.
\newblock \showarticletitle{{Privacy-concerned multiagent planning}}.
\newblock {\em Knowledge and Information Systems\/} {48}, 3 (2016), 581--618.
\newblock
\showISSN{0219-1377}


\bibitem[\protect\citeauthoryear{To{\v{z}}i{\v{c}}ka, {\v{S}}tolba, and
  Komenda}{To{\v{z}}i{\v{c}}ka et~al\mbox{.}}{2017}]%
        {Tozicka17}
{Jan To{\v{z}}i{\v{c}}ka}, {Michal {\v{S}}tolba}, {and} {Anton{\'{\i}}n
  Komenda}. 2017.
\newblock \showarticletitle{The Limits of Strong Privacy Preserving Multi-Agent
  Planning}. In {\em Proceedings of the 27th International Conference on
  Automated Planning and Scheduling (ICAPS)}. 221--229.
\newblock


\bibitem[\protect\citeauthoryear{van~der Krogt}{van~der Krogt}{2007}]%
        {vanderKrogt07b}
{Roman van~der Krogt}. 2007.
\newblock \showarticletitle{Privacy Loss in Classical Multiagent Planning}. In
  {\em Proceedings of the IEEE/WIC/ACM International Conference on Intelligent
  Agent Technology (IAT)}. 168--174.
\newblock


\bibitem[\protect\citeauthoryear{van~der Krogt}{van~der Krogt}{2009}]%
        {Krogt09}
{Roman van~der Krogt}. 2009.
\newblock \showarticletitle{Quantifying privacy in multiagent planning}.
\newblock {\em Multiagent and Grid Systems\/} {5}, 4 (2009), 451--469.
\newblock


\bibitem[\protect\citeauthoryear{Wilkins}{Wilkins}{1988}]%
        {Wilkins88}
{David~E. Wilkins}. 1988.
\newblock {\em Practical Planning: Extending the Classical {AI} Planning
  Paradigm}.
\newblock Morgan Kaufmann.
\newblock


\bibitem[\protect\citeauthoryear{Wilkins and Myers}{Wilkins and Myers}{1998}]%
        {Wilkins98}
{David~E. Wilkins} {and} {Karen~L. Myers}. 1998.
\newblock \showarticletitle{A multiagent planning architecture}. In {\em
  Proceedings of the 4th International Conference on Artificial Intelligence
  Planning Systems (AIPS)}. 154--162.
\newblock


\bibitem[\protect\citeauthoryear{Wolverton and desJardins}{Wolverton and
  desJardins}{1998}]%
        {Wolvertond98}
{Michael Wolverton} {and} {Marie desJardins}. 1998.
\newblock \showarticletitle{Controlling Communication in Distributed Planning
  Using Irrelevance Reasoning}. In {\em Proceedings of the 15th National
  Conference on Artificial Intelligence (AAAI)}. 868--874.
\newblock


\bibitem[\protect\citeauthoryear{Wooldridge}{Wooldridge}{1997}]%
        {Wooldridge97}
{Michael Wooldridge}. 1997.
\newblock \showarticletitle{Agent-Based Software Engineering}.
\newblock {\em IEE Proceedings - Software Engineering\/} {144}, 1 (1997),
  26--37.
\newblock


\bibitem[\protect\citeauthoryear{Zhang and Kambhampati}{Zhang and
  Kambhampati}{2014}]%
        {Zhang14}
{Yu Zhang} {and} {Subbarao Kambhampati}. 2014.
\newblock \showarticletitle{A Formal Analysis of Required Cooperation in
  Multi-agent Planning}.
\newblock {\em CoRR\/}  {abs/1404.5643} (2014).
\newblock
\showURL{%
\url{http://arxiv.org/abs/1404.5643}}


\end{thebibliography}

\received{August 2016}{April 2017}{July 2017}




\end{document}